Article

# Detection of Risk Predictors of COVID-19 Mortality with Classifier Machine Learning Models Operated with Routine Laboratory Biomarkers

**Mehmet Tahir Huyut** [1,\*], **Andrei Velichko** [2] **and Maksim Belyaev** [2]

[1] Department of Biostatistics and Medical Informatics, Faculty of Medicine, Erzincan Binali Yıldırım University, Erzincan 24000, Turkey

[2] Institute of Physics and Technology, Petrozavodsk State University, 33 Lenin Ave., 185910 Petrozavodsk, Russia; velichko@petrsu.ru (A.V.); biomax89@yandex.ru (M.B.)

\* Correspondence: tahir.huyut@erzincan.edu.tr

**Abstract:** Early evaluation of patients who require special care and who have high death-expectancy in COVID-19, and the effective determination of relevant biomarkers on large sample-groups are important to reduce mortality. This study aimed to reveal the routine blood-value predictors of COVID-19 mortality and to determine the lethal-risk levels of these predictors during the disease process. The dataset of the study consists of 38 routine blood-values of 2597 patients who died ($n$ = 233) and those who recovered ($n$ = 2364) from COVID-19 in August–December, 2021. In this study, the histogram-based gradient-boosting (HGB) model was the most successful machine-learning classifier in detecting living and deceased COVID-19 patients (with squared F1 metrics $F1^2 = 1$). The most efficient binary combinations with procalcitonin were obtained with D-dimer, ESR, D-Bil and ferritin. The HGB model operated with these feature pairs correctly detected almost all of the patients who survived and those who died (precision > 0.98, recall > 0.98, $F1^2$ > 0.98). Furthermore, in the HGB model operated with a single feature, the most efficient features were procalcitonin ($F1^2$ = 0.96) and ferritin ($F1^2$ = 0.91). In addition, according to the two-threshold approach, ferritin values between 376.2 μg/L and 396.0 μg/L ($F1^2$ = 0.91) and procalcitonin values between 0.2 μg/L and 5.2 μg/L ($F1^2$ = 0.95) were found to be fatal risk levels for COVID-19. Considering all the results, we suggest that many features combined with these features, especially procalcitonin and ferritin, operated with the HGB model, can be used to achieve very successful results in the classification of those who live, and those who die from COVID-19. Moreover, we strongly recommend that clinicians consider the critical levels we have found for procalcitonin and ferritin properties, to reduce the lethality of the COVID-19 disease.

## 1. Introduction

The complexity of severe acute respiratory syndrome coronavirus 2 (SARS-CoV-2/COVID-19), and the rapid spread of the disease, causing serious and fatal complications has focused researchers' attention on the clinical course of the disease [1,2]. The disease has placed an unprecedented strain on healthcare systems around the world, and has involved medical professionals in an unknown and challenging effort to treat populations of patients with a new and deadly disease [3–5].

Although important information about the genetic structure of this new virus has been obtained [6] and data on the symptoms of the disease are shared in the medical community [7], there are still many severe cases. Mortality rates for this disease differ among countries [8–10], and workload in hospitals affects mortality [5,10].

While mild symptoms (fever, dry cough, shortness of breath, myalgia, fatigue, etc.) are seen in most of the patients with SARS-CoV-2 infection, it has been stated that acute respiratory distress syndrome, septic shock, bleeding, coagulation disorder, and metabolic acidosis can be seen, and result in death in severe cases, noting that this disease can accompany multi-organ dysfunction and cause a variety of symptoms [11–15]. Onur et al.[16] and Huyut et al. [12] reported that COVID-19 disease may be asymptomatic or associated with severe ARDS, thought to be due to an inflammatory cytokine-storm. Furthermore, Chalmers et al. [17] noted that excessive and uncontrolled release of proinflammatory cytokines may be considered the most important primary cause of death from coronavirus, as has been reported in other infections caused by pathogenic coronaviruses. However, attempts to identify and treat hyperinflammation associated with the COVID-19 infection, continue [18].

Many studies have indicated that male gender, advanced age, comorbidities such as chronic obstructive pulmonary disease (COPD) and diabetes mellitus, and some routine laboratory tests such as D-dimer, procalcitonin, and CRP are associated with worse outcomes of the disease [12,15,19–22]. These studies also stated that COVID-19 patients with severe pneumonia have decreased serum-albumin and prealbumin levels, and signs of deterioration in liver and kidney functions.

Most of the previous routine blood-studies were aimed at identifying features that affect the diagnosis and prognosis of COVID-19 [3,22,23]. However, as the number of infected and fatal cases increases worldwide, there remains a need for a detailed investigation of clinical, radiological, and laboratory features, and, more importantly, mortality risk-factors in severe COVID-19 patients [11,24]. Zhang et al. [24] noted that there may be changes in the previously detected predictive-values of mortality in severe and critical COVID-19 patients. Therefore, Ponti et al. [25] and Huyut [3] stated that the determination of effective routine laboratory-biomarkers that can classify COVID-19 patients according to their fatal risk, supported by studies with large samples, is essential in order to guarantee rapid and effective treatment. Indeed, many studies have emphasized that the usefulness and effective breakpoints of many laboratory markers such as ferritin and D-dimer in predicting COVID-19 mortality have not been fully determined [11,17,26–28]. Therefore, Chalmers et al. [17] and Cheng et al. [29] stated that the predictive role of routine laboratory-features in identifying risk factors affecting COVID-19 mortality needs further confirmation.

However, it is known that even the most knowledgeable and experienced physicians can interpret little of the information contained in routine blood laboratory-results, and it is extremely difficult to determine the severity of COVID-19 patients based on laboratory findings alone [30]. In contrast, machine learning (ML) models have been successfully used to recognize subtle patterns in data to distinguish latent-association patterns between routine blood-parameters and disease [3,13,22,31,32]. Indeed, several studies have shown that ML models can predict COVID-19 patient groups with high accuracy, using patient demographics, physiological characteristics, and RBV data [33–36].

In our previous studies, we determined routine blood-values that predict the diagnosis and prognosis of COVID-19 with various supervised ML models and LogNNet [3,4,13,22,32]. Huyut and Velichko [22] used only three RBV features with the LogNNet model to predict the diagnosis and prognosis of COVID-19 disease. They achieved an accuracy of 99.17% in the diagnosis of the disease and an accuracy of 82.7% in determining the prognosis of the disease. Velichko et al. [32] achieved 100% accuracy in the diagnosis of COVID-19, using 11 RBV features with a histogram-based gradient-boosting model in their study. Huyut [3] classified severe and mild COVID patients from a large patient population, using 28 RBV features, and the models with the highest classification accuracy were locally weighted learning (97.86%) and k-nearest neighbor (94.05%).

Huyut and İlkbahar [4] used various biomarkers with the CHAID decision tree to detect the diagnosis and prognosis of COVID-19. The model showed 81.6% accuracy in recognizing the disease and 93.5% accuracy in determining the prognosis of the disease.

Formica et al. [37] developed an ML model for early diagnosis of disease, using eight RBV features, and reported an 82% specificity with 83% sensitivity; however, the analysis was based on a small sample (171 patients). Banerjee et al. [38] classified a patient cohort of 598 cases, 39 of whom were COVID-19 positive, by various ML methods, using 12 RBV features, and reported good specificity (91%) but very low sensitivity (43%). Avila et al.[39] developed a Bayesian model using the dataset of 12 RBV features, and reported a sensitivity and specificity of 76.7% in the diagnosis of the disease. Joshi et al. [40] developed a trained logistic-regression model using only hemogram data on a dataset of 380 cases, and reported 93% sensitivity but low 43% specificity in the diagnosis of COVID-19 disease. Zhu et al. [41] used 78 features, consisting of demographic, clinical, and RBV values, with a deep neural network model to predict the mortality of COVID-19. They found the success of the method in diagnosis to be 95.4% AUC. Soltan et al. [42] ran models of multivariate logistic regression, random forests and extreme gradient-supported trees on more than 50 RBV data, to identify COVID-19. The most successful model in the diagnosis of the disease was the XGBoost method, with 85% sensitivity and 90% accuracy. Soares [43] developed an ML model using 15 RBV parameters to diagnose COVID-19 on a sample of 599 people, 81 of whom were COVID-19-positive. Combining the SVM, ensemble, and SMOTE Boost models, this model had 86% specificity and 70% success in diagnosing the disease.

In this study, 34 routine blood-values were determined with a statistical approach, in order to find the most successful ML classifier-model in detecting patients with COVID-19 who lived and died. Therefore, 16 ML models were run with these features, and the most successful model (histogram-based gradient-boosting/HGB) was determined. The predictors of mortality of COVID-19 disease were revealed with the HGB model. The performance of individual and binary combinations of these predictors in detecting patient groups was obtained with the HGB model. In addition, correlations between patient groups and binary combinations of features were interpreted in detail. In addition, cut-off values for these characteristics (mortality risk-levels) were calculated using one- and two-threshold approaches in the classification of patient groups according to direct characteristics. We think that the findings of this study will be an important motivational tool for clinicians in estimating the mortality of COVID-19 and detecting severe patients.

The paper has the following structure. Section 2 describes the data-collection procedure, metrics, characteristics of the participants, the feature-selection procedure, the one- and two-threshold approaches for classification based on the values of direct features and a new $F1^2$ criterion for the classification metric. Section 3 describes the correlations of features with patient groups, statistical differences of features between groups, classification results according to ML models, classification results of individual and pairwise combinations of features operated with the HGB model, and classification results according to one- and two-threshold values of the features. Section 4 discusses the results and compares them with known developments. Section 5 presents the limitations of the study. Finally, in Section 6, a general description of the study and its scientific significance is given.

## 2. Materials and Methods

In this retrospective cohort study, data suitable for our criteria were collected from the Erzincan Binali Yıldırım University Mengücek Gazi Training and Research Hospital information system between August and November 2021, and included in the study. The laboratory data of the patients were the routine blood-values measured at the time of admission to the hospital. The information about the patients was followed up until exit, and exit information was recorded. In our hospital, a diagnosis of SARS-CoV-2 was made using real-time reverse transcription polymerase chain reaction (RT-PCR) only, on nasopharyngeal or oropharyngeal swabs.

## 2.1. Measurements

Sysmex XN-1000 Hematology System (Sysmex Corporation, Kobe, Japan) was used to carry out cell blood count. Biochemical tests were analyzed by the spectrophoto spectrophotometric method using Beckman Coulter Olympus AU2700 Plus Chemistry Analyzer (Beckman Coulter, Tokyo, Japan) from serum Prothrombin time (PT), activated partial prothrombin time (aPTT), and fibrinogen were determined with a digital coagulation device from Ceveron-Alpha (Diapharma Group Inc., West Chester, Canada). The erythrocyte sedimentation rate (ESR) was measured using the TEST 1 BCL instrument (Alifax, Polverara, Italy), based on the principle of photometric capillary-flow kinetic analysis. Ferritin was evaluated with a chemiluminescence immunoassay (Centaur XP, Siemens Healthcare, Germany). C-reactive protein (CRP) was measured using the nephelometric method in the BNTM II System (Siemens, Munich, Germany). Procalcitonin (PCT), D-dimer, and troponin were analyzed from whole blood on the AQT90 flex RadiometerVR (Bronshoj, Denmark). All patient data were double-checked and analyzed by the research team.

## 2.2. Characteristics of Participants and Defined Datasets

In this study, only RBV data (features) of 2597 patients who were diagnosed with COVID-19 and treated at the hospital during the specified dates were used. During the treatment period, 233 (9.0%) of these patients died, while 2364 (91.0%) survived. Of the patients who lost their lives, 143 (61.3%) were male, while 90 (38.7%) were female. The mean age of the surviving patients was 55 years, while the mean age of the deceased patients was 76 years.

The routine laboratory-information of these patients was examined. The RBVs (features) that were measured from at least 80% of the patients were used. Missing data in this study were completed with the mean of the relevant parameter distribution, and outliers were normalized. A total of 38 routine blood-values calibrated from approximately 70 parameters were used in this study. The data used in this study will be used as "SARS-CoV-2-RBV3" (Supplementary Materials). The SARS-CoV-2-RBV3 dataset includes immunological, hematological, and biochemical parameters (Table 1).

**Table 1.** Feature-numbering for SARS-CoV-2-RBV3 dataset.

| № | Feature | № | Feature | № | Feature | № | Feature |
|---|---|---|---|---|---|---|---|
| 1 | ALT | 11 | LDH | 21 | MCV | 31 | Ferritin |
| 2 | AST | 12 | eGFR | 22 | MONO | 32 | Fibrinogen |
| 3 | Albumin | 13 | UA | 23 | MPV | 33 | INR |
| 4 | ALP | 14 | BASO | 24 | NEU | 34 | PT |
| 5 | Amylase | 15 | EOS | 25 | PLT | 35 | PCT |
| 6 | CK-MB | 16 | HCT | 26 | RBC | 36 | ESR |
| 7 | D-Bil | 17 | HGB | 27 | RDW | 37 | Troponin |
| 8 | Glucose | 18 | LYM | 28 | WBC | 38 | aPTT |
| 9 | Creatinine | 19 | MCH | 29 | CRP | | |
| 10 | CK | 20 | MCHC | 30 | D-dimer | | |

ALT: alanine aminotransaminase; AST: aspartate aminotransferase; ALP: alkaline phosphatase; CK-MB: creatine kinase myocardial band; D-Bil: direct bilirubin; CK: creatinine kinase; LDH: lactate dehydrogenase; eGFR; estimated glomerular filtration rate; UA: uric acid; BASO: basophil count; EOS: eosinophil count; HCT: hematocrit; HGB: hemoglobin; LYM: lymphocyte count; MCH: mean corpuscular hemoglobin; MCHC: mean corpuscular hemoglobin concentration; MCV: mean corpuscular volume; MONO: monocyte count; MPV: mean platelet volume; NEU: neutrophil count; PLT: platelet count; RBC: red blood cells; RDW: red cell distribution width; WBC: white blood cell count; CRP: C-reactive protein; INR: international normalized ratio; PT: prothrombin time; PCT: procalcitonin; ESR: erythrocyte sedimentation rate; aPTT: activated partial prothrombin time.

These patients were of Turkish and Kurdish ethnicity. Only data from individuals over the age of 18 were recorded. Since it is a retrospective study, comorbidity data of the patients could not be obtained. In the SARS-CoV-2-RBV3 dataset, surviving patients were coded as 0, and patients who died were coded as 1 (survived COVID-19 = 0, non-survived COVID-19 = 1).

The features in this dataset are calibrated, and include almost all of the RBV values that are the subject of studies on COVID-19 mortality. Therefore, we think that the bias of our study using this data set was minimized in comparison with the literature. In addition, the use of our data set, which we can share upon request from researchers, is important in terms of demonstrating the reproducibility and suitability of the results.

*2.3. Feature Selection for ML Models with Statistical Approach*

In order to evaluate the difference of 38 RBV values between patients who survived COVID-19 and those who died, the assumptions of the parametric tests were checked first. The assumption of normality was analyzed with the Shapiro–Wilk test, and the homogeneity of variances in the groups was analyzed with Levene's test. Since the assumptions of the parametric tests were not met, the significance of the difference of 38 features between the patient groups (two independent groups) was analyzed using the Mann–Whitney U test [44,45] and p-values were calculated. A total of 34 features were judged to be statistically different between patient groups, and these features were used by ML models for classification. It is understood that the features selected with this approach may be the determining factors between patients dying and patients surviving. In addition, we used features that were statistically different between patient groups as inputs to the ML models. This approach increased the clinical reliability of our results and reduced biased results.

*2.4. Threshold Approach*

The simplest approach for classification by one feature in the presence of only two classes is based on determining the threshold values separating the classes $V_{th}$ [22].

2.4.1. One-Threshold Approach

For a one-threshold approach for the SARS-CoV-2-RBV3 dataset, we introduce the threshold value Type 1 or Type 2, in accordance with the rule:

$$\begin{cases} \text{Type 1:} & \text{if feature value} \geq V_{th} \text{ then "}survived\text{" else "}non-survived\ COVID-19\text{"} \\ \text{Type 2:} & \text{if feature value} \geq V_{th} \text{ then "}non-survived\text{" else "}survived\ COVID-19\text{"} \end{cases} \quad (1)$$

The threshold type indicates which side of the threshold the non-survived and survived classes are on.

2.4.2. Two-Threshold Approach

For a two-threshold approach for the SARS-CoV-2-RBV3 dataset, we introduce the threshold value Type 1 or Type 2, in accordance with the rule:

$$\begin{cases} \text{Type 1:} & \text{if feature (value} \geq V_{th\_1}) and (\text{value} \leq V_{th\_2}) \text{ then "}survived\text{" else "}non-survived\ COVID-19\text{"} \\ \text{Type 2:} & \text{if feature (value} \geq V_{th\_1}) and (\text{value} \leq V_{th\_2}) \text{ then "}non-survived\text{" else "}survived\ COVID-19\text{"} \end{cases} \quad (2)$$

The main metrics were calculated after balancing the dataset. The k-fold validation was not used when calculating $A_{th}$ and $F1^2$. The threshold values $V_{th}$, $V_{th\_1}$, $V_{th\_2}$ were determined by stepwise enumeration and by finding the maximum value of $A_{th}$.

*2.5. $F1^2$ Metric*

To select the most significant features, we introduced an additional metric, $F1^2$, equal to the product of the F1 metrics of the two classes.

$$F1^2 = F1(non-survived\ COVID-19) \cdot F1(survived\ COVID-19) \quad (3)$$

## 3. Results

### 3.1. Correlation Analysis of Dataset SARS-CoV-2-RBV3

Figure 1 shows the results of the correlation analysis of the diagnosis of features using the three types of Pearson, Spearman, and Kendall correlations over the entire volume of the SARS-CoV-2-RBV3 database. It can be seen that the Spearman and Kendall correlations have very similar values. The Pearson correlation gives in general a smaller number of features that correlate with the diagnosis, so we will use the Spearman correlation as the main one.

Full Spearman heatmaps across the entire database and by class (survived COVID-19 and non-survived COVID-19) are shown in Figure 2.

Figure 2b,c show that the non-survived COVID-19 class is characterized by an increased correlation between features, compared with the survived COVID-19 class, which indicates poor self-regulation in the body.

The most significant changes in the correlation of features of the non-survived COVID-19 class compared with the survived COVID-19 class are presented in Table 2. Here, the qualitative change is denoted as 'Down' and 'Up'. For some pairs of features, the correlation increased, while for some it fell.

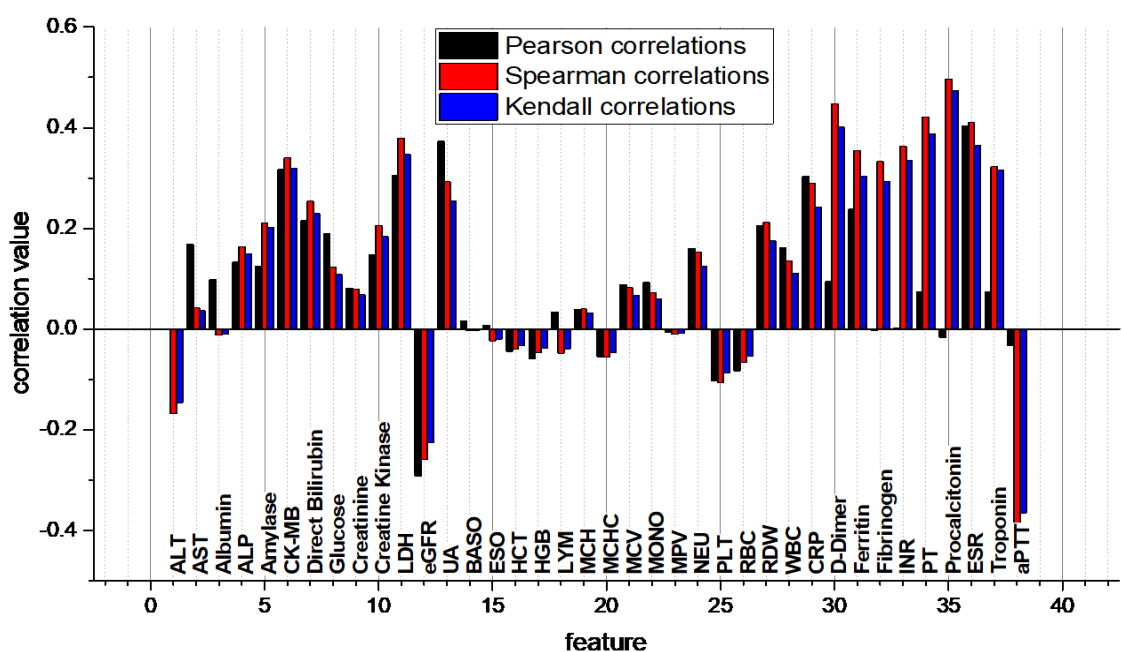

**Figure 1.** Pearson, Spearman and Kendall correlations of the SARS-CoV-2-RBV3 dataset for COVID-19 mortality-feature pairs.

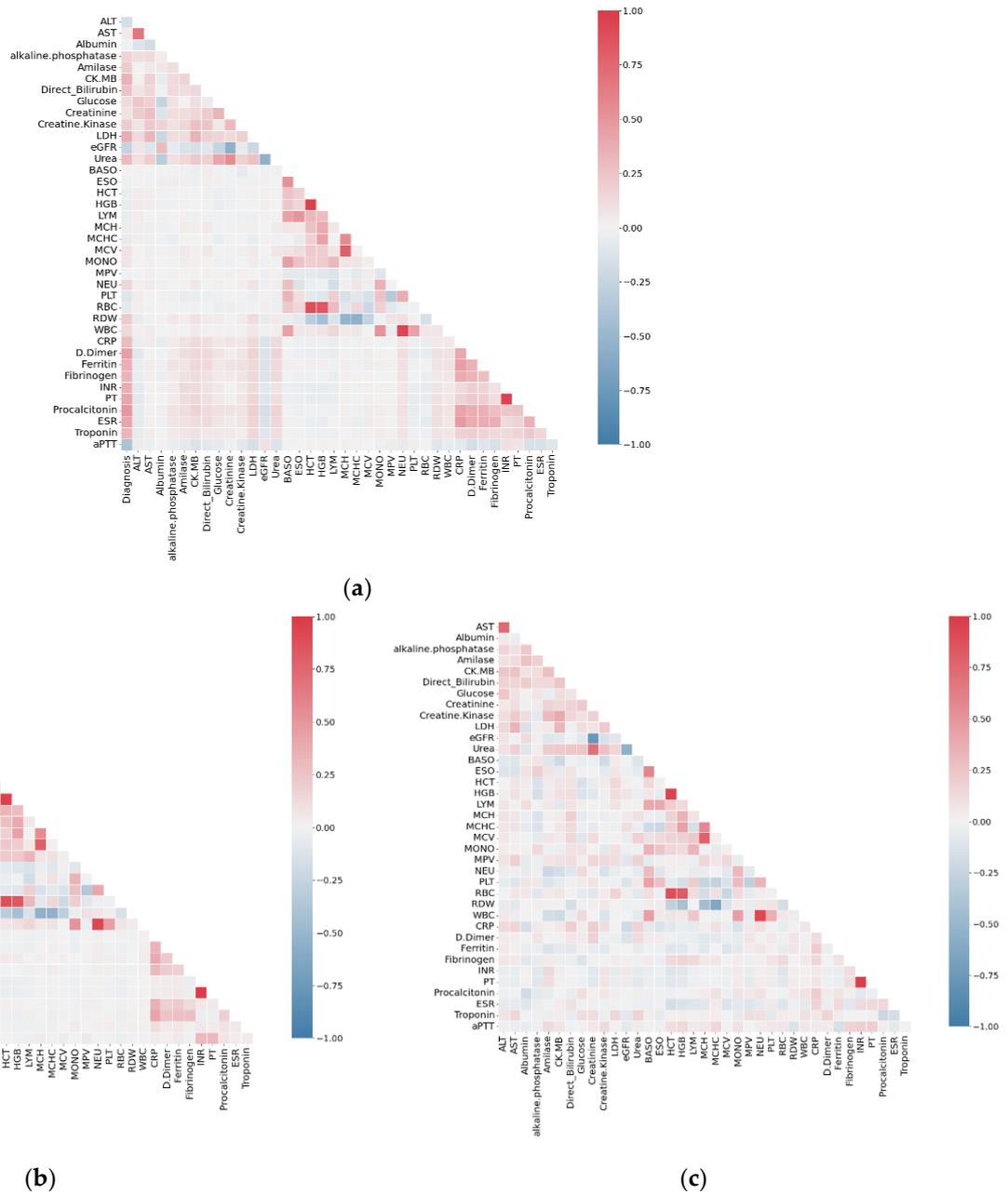

**Figure 2.** Spearman correlation analysis results for (**a**) the entire database, (**b**) survived COVID-19 class, and (**c**) non-survived COVID-19 class from the SARS-CoV-2-RBV3 dataset.

**Table 2.** Changes in the correlation of feature-pairs of the non-survived COVID-19-class compared with the survived COVID-19-class.

| № | № | Spearman Survived COVID-19 | Spearman Non-Survived COVID-19 | Change in the Correlation of Features, in Present of Non-Survived COVID-19 | Feature | Feature |
|---|---|---|---|---|---|---|
| 8 | 3 | −0.31194 | 0.01274 | Down | Glucose | Albumin |
| 13 | 3 | −0.3753 | −0.10287 | Down | UA | Albumin |
| 36 | 29 | 0.42911 | 0.16647 | Down | ESR | CRP |
| 10 | 5 | 0.05417 | 0.30605 | Up | CK | Amylase |
| 12 | 9 | −0.53482 | −0.77476 | Up | eGFR | Creatinine |
| 5 | 3 | −0.03084 | 0.26142 | Up | Amylase | Albumin |
| 12 | 3 | 0.3479 | 0.11987 | Down | eGFR | Albumin |
| 30 | 29 | 0.3365 | 0.1146 | Down | D-dimer | CRP |
| 6 | 5 | 0.05413 | 0.26867 | Up | CK-MB | Amylase |
| 32 | 30 | 0.21431 | $-1.48572 \times 10^{-4}$ | Down | Fibrinogen | D-dimer |
| 10 | 6 | 0.18243 | 0.39221 | Up | CK | CK-MB |
| 36 | 30 | 0.25085 | −0.04489 | Down | ESR | D-dimer |
| 24 | 5 | 0.00641 | −0.21234 | Up | NEU | Amylase |
| 14 | 6 | −0.00966 | −0.20355 | Up | BASO | CK-MB |
| 4 | 3 | −0.03222 | 0.22455 | Up | ALT | Albumin |
| 26 | 18 | 0.30758 | 0.1153 | Down | RBC | LYM |
| 9 | 1 | 0.24486 | 0.05265 | Down | Creatinine | ALT |
| 16 | 15 | 0.20104 | 0.01024 | Down | HCT | EOS |
| 9 | 2 | 0.30293 | 0.1127 | Down | Creatinine | AST |
| 17 | 14 | 0.24856 | 0.05841 | Down | HGB | BASO |
| 25 | 15 | 0.0849 | 0.27473 | Up | PLT | EOS |
| 25 | 20 | −0.08083 | −0.27021 | Up | PLT | MCHC |
| 7 | 3 | −0.0119 | 0.19865 | Up | D-Bil | Albumin |
| 20 | 15 | −0.04499 | −0.23071 | Up | MCHC | EOS |
| 31 | 29 | 0.3931 | 0.2085 | Down | Ferritin | CRP |
| 28 | 6 | 0.0179 | −0.20122 | Up | WBC | CK-MB |
| 27 | 21 | −0.30218 | −0.11908 | Down | RDW | MCV |
| 7 | 6 | 0.0635 | 0.24642 | Up | D-Bil | CK-MB |
| 36 | 32 | 0.27287 | −0.09309 | Down | ESR | Fibrinogen |
| 20 | 14 | 0.02204 | −0.1999 | Up | MCHC | BASO |
| 13 | 8 | 0.42328 | 0.24648 | Down | UA | Glucose |
| 15 | 4 | 0.01565 | 0.19167 | Up | EOS | ALT |
| 31 | 30 | 0.22095 | −0.04543 | Down | Ferritin | D-dimer |
| 6 | 1 | 0.06557 | 0.23993 | Up | CK-MB | ALT |
| 32 | 29 | 0.31919 | 0.14655 | Down | Fibrinogen | CRP |
| 23 | 2 | 0.00857 | 0.18043 | Up | MPV | AST |
| 3 | 2 | −0.20943 | −0.03802 | Down | Albumin | AST |
| 35 | 3 | −0.01609 | −0.18485 | Up | PCT | Albumin |
| 30 | 9 | −0.00743 | 0.17441 | Up | D-dimer | Creatinine |
| 23 | 13 | 0.00586 | 0.1727 | Up | MPV | UA |

*3.2. Comparison of RBV Features of Surviving and Non-Surviving COVID-19 Patients and Comparison of ML Classifiers*

The statistical comparison results of 38 characteristics of surviving and non-surviving COVID-19 patients are presented in Table 3. Except for albumin, BASO, EOS, and MPV, the other 34 features were judged as statistically different between the patient groups. The 34 features selected here were used as inputs to identify patient groups with ML models and the classification performance of the models was obtained (Table 4 and Figure 3). Considering the $F1^2$ (see Equation (3)) criterion derived from the F1 metrics of the classes in the classification of patient groups, it was found that the most successful model was HGB ($F1^2$ value: 1). After HGB, the most successful models were Adaboost, Extra Trees, KNN, RF, and SVM-LK, (at least $F1^2 > 0.99$ in these models). The most unsuccessful model was quadratic discriminant analysis ($F1^2$ value: 0.72).

Table 3. Descriptive statistics of RBV values of surviving and non-surviving COVID-19 groups.

| Parameters (Units) | Surviving Group | | | Non-Surviving Group | | | p |
|---|---|---|---|---|---|---|---|
| | Median | Percentile 25 | Percentile 75 | Median | Percentile 25 | Percentile 75 | |
| ALT (U/L) | 35.31 | 24.00 | 35.31 | 23.00 | 15.00 | 35.20 | <0.001 |
| AST (U/L) | 33.24 | 25.00 | 33.24 | 32.00 | 22.00 | 47.23 | 0.033 |
| Albumin (g/L) | 38.59 | 38.59 | 38.59 | 38.29 | 33.00 | 43.54 | 0.539 |
| ALP (U/L) | 84.10 | 84.10 | 84.10 | 103.23 | 72.00 | 103.23 | <0.001 |
| Amylase (U/L) | 73.70 | 73.70 | 73.70 | 101.00 | 58.00 | 107.62 | <0.001 |
| CK-MB (U/L) | 18.79 | 18.79 | 18.79 | 32.75 | 19.40 | 32.75 | <0.001 |
| D-Bil. (mg/dL) | 0.13 | 0.13 | 0.13 | 0.25 | 0.12 | 0.27 | <0.001 |
| Glucose (mg/dL) | 136.03 | 108.00 | 136.03 | 145.00 | 113.00 | 188.00 | <0.001 |
| Creatinine (mg/dL) | 1.14 | 0.90 | 1.14 | 1.11 | 0.86 | 1.64 | <0.001 |
| CK (U/L) | 104.26 | 83.00 | 104.26 | 220.00 | 79.00 | 350.53 | <0.001 |
| LDH (U/L) | 252.94 | 252.94 | 252.94 | 309.76 | 309.76 | 309.76 | <0.001 |
| eGFR | 82.74 | 82.74 | 85.10 | 62.16 | 44.47 | 82.50 | <0.001 |
| UA (mg/dL) | 38.80 | 32.00 | 38.80 | 56.74 | 39.13 | 75.95 | <0.001 |
| BASO ($10^3/\mu L$) | 0.02 | 0.01 | 0.04 | 0.021 | 0.014 | 0.044 | 0.869 |
| EOS ($10^3/\mu L$) | 0.04 | 0.01 | 0.12 | 0.03 | 0.00 | 0.12 | 0.232 |
| HCT (%) | 39.55 | 36.00 | 43.20 | 38.80 | 34.90 | 42.30 | 0.041 |
| HGB (g/L) | 13.30 | 12.00 | 14.65 | 13.10 | 11.50 | 14.50 | 0.016 |
| LYM ($10^3/\mu L$) | 1.46 | 0.99 | 2.03 | 1.32 | 0.85 | 1.88 | 0.015 |
| MCH (pg) | 28.60 | 27.30 | 29.60 | 28.80 | 27.20 | 30.10 | 0.041 |
| MCHC (g/dL) | 33.80 | 32.90 | 34.70 | 33.50 | 32.40 | 34.60 | 0.004 |
| MCV (fL) | 83.90 | 80.80 | 87.00 | 85.20 | 81.80 | 88.90 | <0.001 |
| MONO ($10^3/\mu L$) | 0.51 | 0.38 | 0.67 | 0.56 | 0.44 | 0.72 | <0.001 |
| MPV (fL) | 10.30 | 9.70 | 10.90 | 10.30 | 9.60 | 11.00 | 0.604 |
| NEU ($10^3/\mu L$) | 4.05 | 2.85 | 5.85 | 5.25 | 3.98 | 7.65 | <0.001 |
| PLT ($10^3/\mu L$) | 229.00 | 184.00 | 287.00 | 200.00 | 166.00 | 250.00 | <0.001 |
| RBC ($10^6/\mu L$) | 4.74 | 4.36 | 5.14 | 4.64 | 4.16 | 4.98 | 0.001 |
| RDW (%) | 13.10 | 12.50 | 13.90 | 14.00 | 13.20 | 15.40 | <0.001 |
| WBC ($10^3/\mu L$) | 6.50 | 5.00 | 8.30 | 7.80 | 6.20 | 10.10 | <0.001 |
| CRP (mg/L) | 6.76 | 3.02 | 23.50 | 72.00 | 17.10 | 72.00 | <0.001 |
| D-dimer (µg/L) | 441.00 | 441.00 | 441.00 | 1277.00 | 1277.00 | 1277.00 | <0.001 |
| Ferritin (µg/L) | 125.95 | 90.90 | 175.80 | 395.00 | 395.00 | 395.00 | <0.001 |
| Fibrinogen (mg/dL) | 321.10 | 321.10 | 321.10 | 350.00 | 350.00 | 350.00 | <0.001 |
| INR | 1.10 | 1.10 | 1.10 | 1.20 | 1.20 | 1.20 | <0.001 |
| PT (Sec) | 13.10 | 13.10 | 13.10 | 14.20 | 14.20 | 14.20 | <0.001 |
| PCT (ng/mL) | 0.12 | 0.12 | 0.12 | 2.75 | 2.53 | 2.75 | <0.001 |
| ESR (nm/hr) | 17.00 | 17.00 | 17.00 | 49.00 | 49.00 | 49.00 | <0.001 |
| Troponin (ng/L) | 16.12 | 10.00 | 19.00 | 53.27 | 15.00 | 75.00 | <0.001 |
| aPTT (Sec) | 32.75 | 32.75 | 32.75 | 32.00 | 32.00 | 32.00 | <0.001 |

$p < 0.05$ was considered significant.

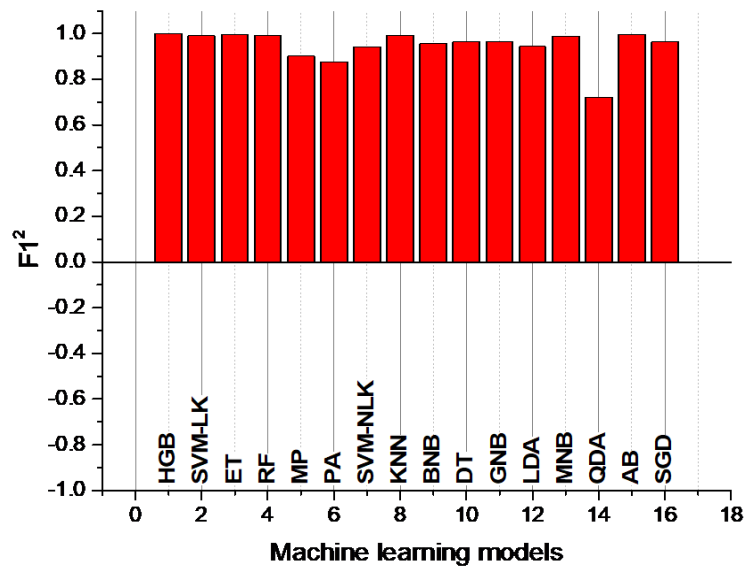

**Figure 3.** Performance of ML models in classifying surviving and non-surviving COVID-19 patients, using the 34 features.

**Table 4.** Classification performance of ML models run with 34 features to detect patient groups.

| No | ML Models | F1$^2$ |
|---|---|---|
| 1 | Histogram-based Gradient Boosting (HGB) | 1.0000 |
| 2 | Adaboost (AB) | 0.9952 |
| 3 | Extra Trees (ET) | 0.9952 |
| 4 | K-nearest neighbors (KNN) | 0.9929 |
| 5 | Random Forest (RF) | 0.9928 |
| 6 | Support Vector Machine with Linear Kernel (SVM-LK) | 0.9904 |
| 7 | Multinomial Naive Bayes (MNB) | 0.9881 |
| 8 | Gaussian Naive Bayes (GNB) | 0.9646 |
| 9 | Stochastic Gradient Descent (SGD) | 0.9642 |
| 10 | Decision Tree (DT) | 0.9642 |
| 11 | Bernoulli Naive Bayes (BNB) | 0.9563 |
| 12 | Linear discriminant analysis (LDA) | 0.9431 |
| 13 | Support Vector Machine with non-linear Kernel (SVM-NLK) | 0.9428 |
| 14 | Multilayer Perceptron (MP) | 0.9011 |
| 15 | Passive-Aggressive (PA) | 0.8772 |
| 16 | Quadratic Discriminant Analysis (QDA) | 0.7212 |

*3.3. Investigation of the Effectiveness of the Models Operating on the One-Feature HGB Model*

It is known that the use of the F1 score may be more useful than the accuracy value in cases where the data distributions are not equal. Figure 4 shows a comparison of F1 metrics for the survived-COVID-19 class, calculated for the original and SMOTE-balanced datasets. It can be seen that, for survived-COVID-19, there is a high F1 value for all features, for both databases.

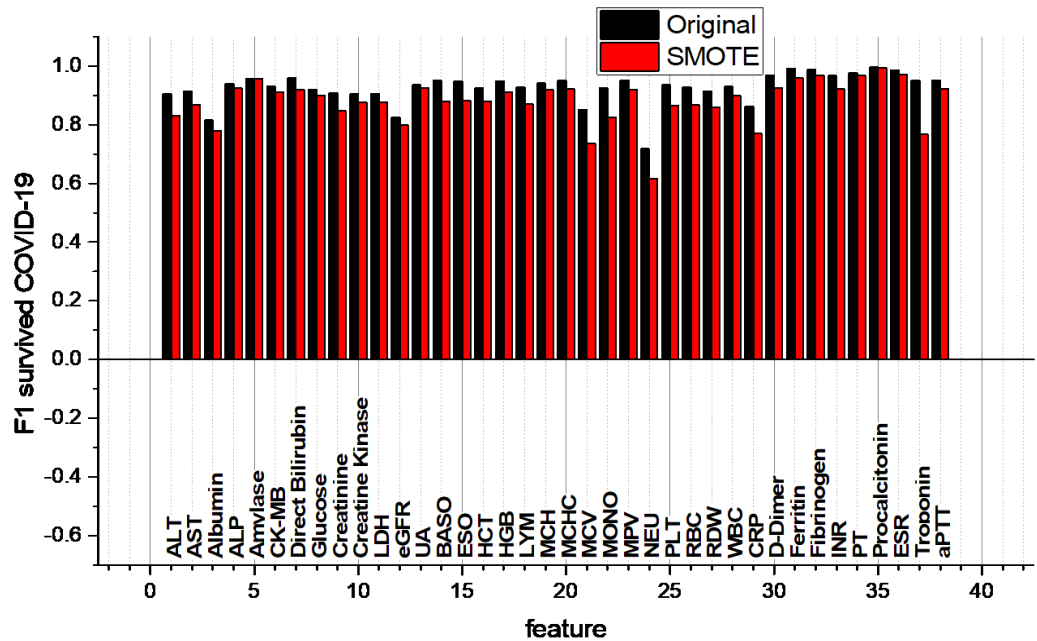

**Figure 4.** F1 metrics for survived-COVID-19 class, calculated for original and SMOTE-balanced datasets.

Figure 5 shows a comparison of F1 metrics for the non-survived-COVID-19 class calculated for the original and SMOTE-balanced datasets. It can be seen that, for non-survived-COVID-19, a high F1 value is observed for most of the features for SMOTE-balanced dataset, while for the original dataset, only some of the features have a high F1. Thus, to select the main features, it is logical to use the results of calculating the metrics for the original dataset. Synthetic data, although well approximated by the model, nevertheless does not allow us to judge the performance of the model with real data. Table A1 in the Appendix A presents the classification result of SARS-CoV-2-RBV3 dataset for the HGB model, using a single input-feature for original dataset, indicating the main classification metrics (Precision, Recall, F1).

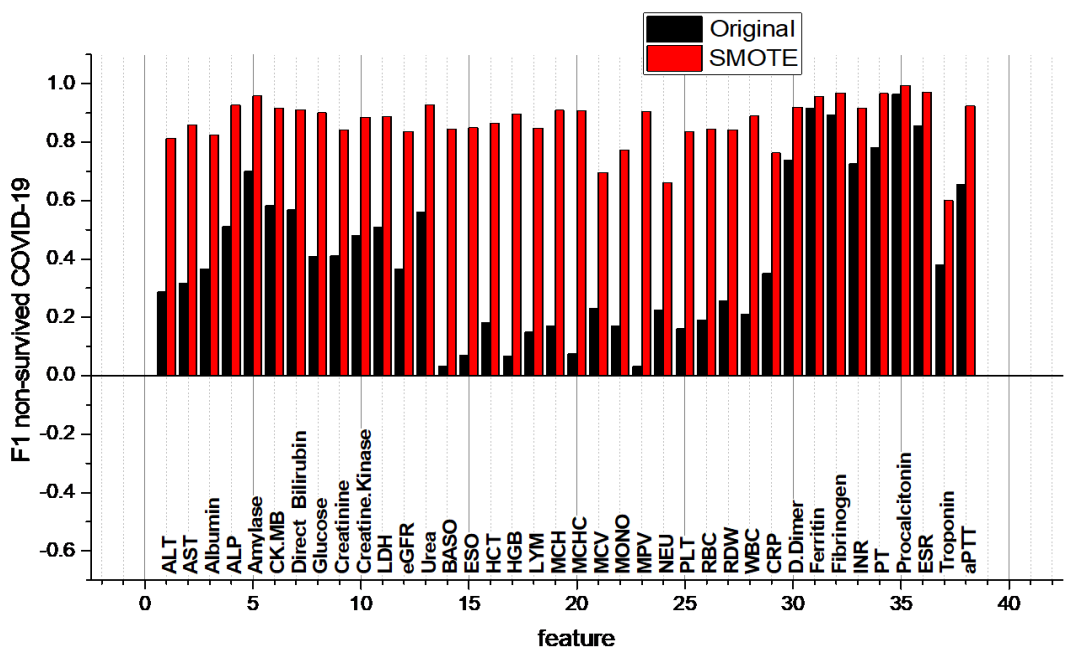

**Figure 5.** F1 metric for non-survived-COVID-19 class, calculated for original and SMOTE-balanced datasets.

## 3.4. $F1^2$ Metric in the Detection of Patient Groups with the HGB Model, One-Threshold, and Two-Threshold Approaches

Figure 6 shows the dependence of $F1^2$ on the feature. Let us agree to consider as the most significant features those features in which $F1^2 \geq 0.5$; this threshold is visualized in the figure by the blue line.

As a result, we obtain a list of the 12 most significant single features for HGB classification, shown in Table 5., in which $F1^2 \geq 0.5$. No high $F1^2$ value was found in the classification of patient groups with the one-threshold-value approach. However, high $F1^2$ values were found in the classification of patient groups with the HGB model operated with a single feature and the two-threshold-value approach. Accordingly, PCT and ferritin properties were found to be the most effective in classification, according to the HGB model operated with one feature and the two-threshold approach.

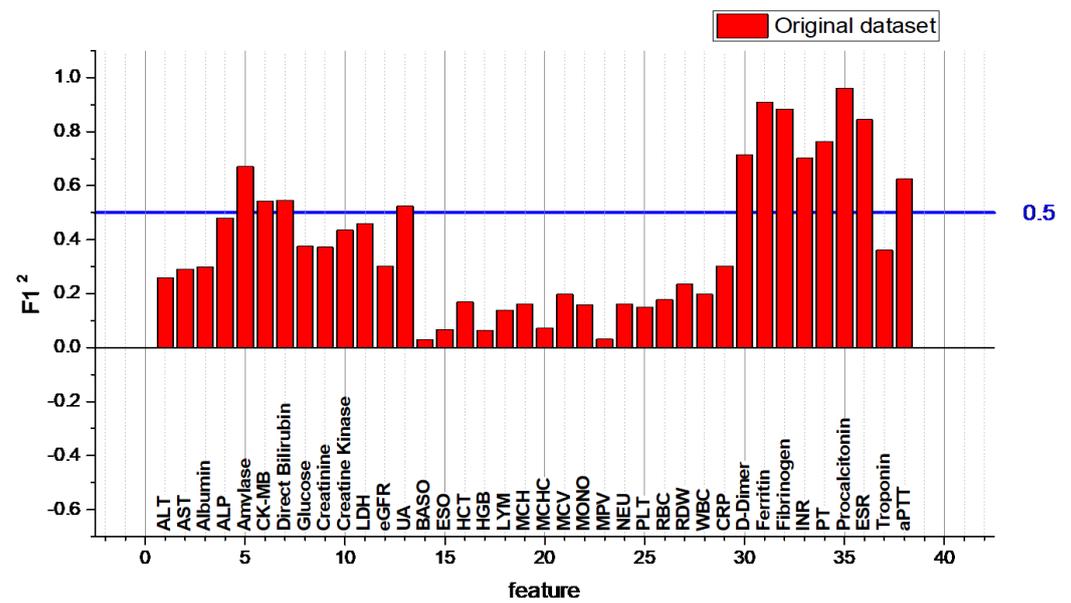

**Figure 6.** $F1^2$ metric of the HGB model according to each feature for the detection of surviving and non-surviving COVID-19 patients.

**Table 5.** List of the 12 most significant single features for classification using the HGB algorithm, with $F1^2$ metric.

| Feature Name | № | $F1^2$ HGB Model | $F1^2$ One-Threshold Approach | $F1^2$ Two-Threshold Approach |
|---|---|---|---|---|
| PCT | 35 | 0.9621 | 0.54277 | 0.95118 |
| Ferritin | 31 | 0.90966 | 0.53731 | 0.90577 |
| Fibrinogen | 32 | 0.88417 | 0.4635 | 0.67443 |
| ESR | 36 | 0.845 | 0.54522 | 0.69842 |
| PT | 34 | 0.76401 | 0.579 | 0.58245 |
| D-dimer | 30 | 0.71535 | 0.6408 | 0.65008 |
| INR | 33 | 0.70204 | 0.58302 | 0.58743 |
| Amylase | 5 | 0.6699 | 0.61374 | 0.6599 |
| aPTT | 38 | 0.62451 | 0.53117 | 0.53603 |
| D-Bil | 7 | 0.54567 | 0.41042 | 0.4068 |
| CK-MB | 6 | 0.54277 | 0.6026 | 0.46247 |
| UA | 13 | 0.52454 | 0.38088 | 0.38088 |

### 3.4.1. Threshold Approach

For the one-threshold approach, we obtained the distribution of the $F1^2$ metric shown in Figure 7. Model types are marked with color (Type 1, Type 2). The complete collection of metrics (Type, $V_{th}$, $A_{th}$, Precision, Recall, F1, $F1^2$) is presented in Table A2 of the Appendix.

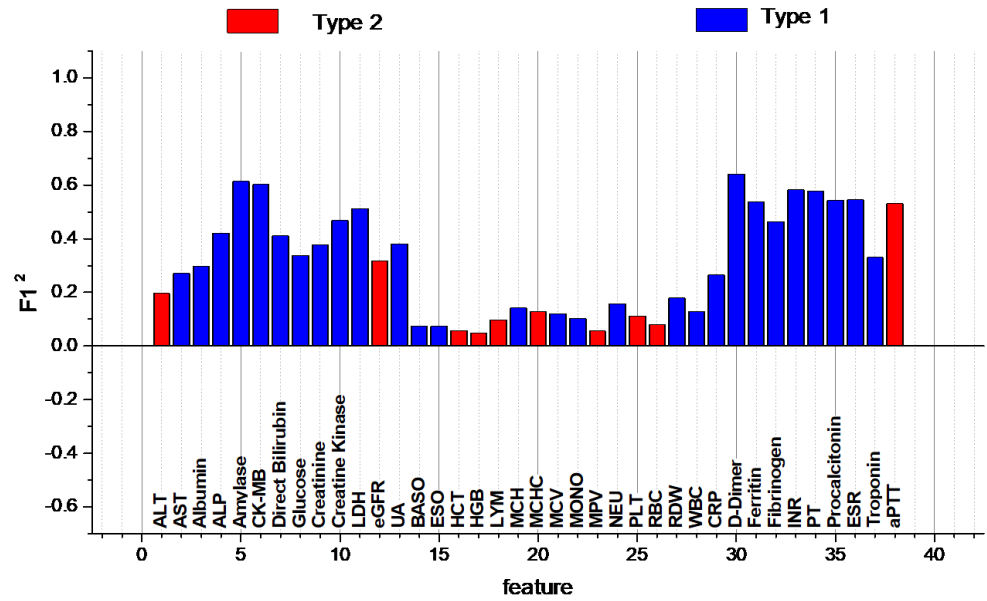

**Figure 7.** The $F1^2$ metric for the classification of surviving and non-surviving COVID-19 patients, according to a single feature for the one-threshold approach, with dependency-type visualization (Type 1, Type 2).

For the two-threshold approach, we obtained the distribution of the $F1^2$ metric shown in Figure 8. Model types are marked with color (Type 1, Type 2). The complete collection of metrics (Type, $V_{th\_1}$, $V_{th\_2}$, $A_{th}$, Precision, Recall, F1, $F1^2$) is presented in Table A3 of the Appendix.

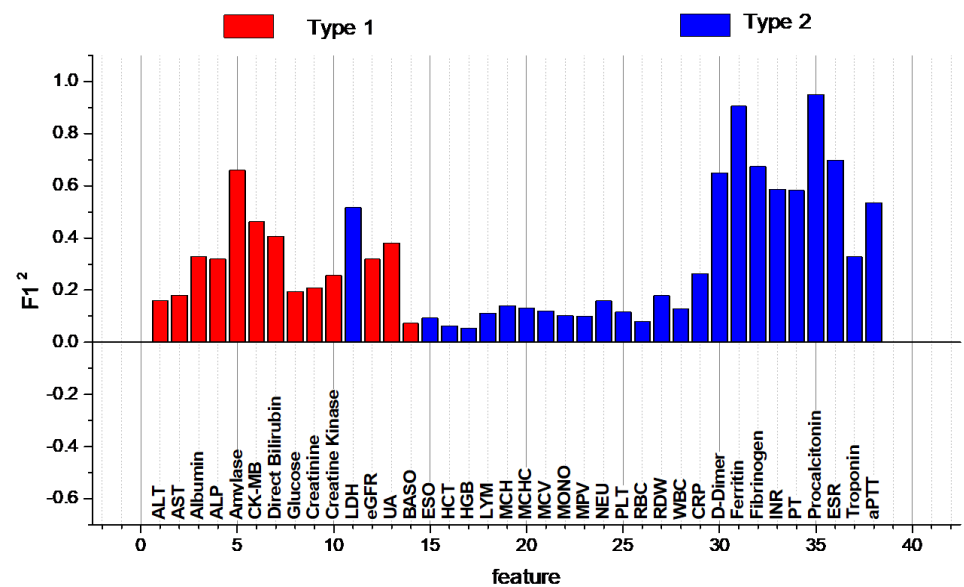

**Figure 8.** The $F1^2$ metric for the classification of surviving and non-surviving COVID-19 patients, according to a single feature for the two-threshold approach, with dependency-type visualization (Type 1, Type 2).

Procalcitonin, ferritin, and fibrinogen samples for the histogram distributions and classification results of the characteristics of surviving and non-surviving COVID-19 patients according to the one-threshold approach, are shown in Figure 9.

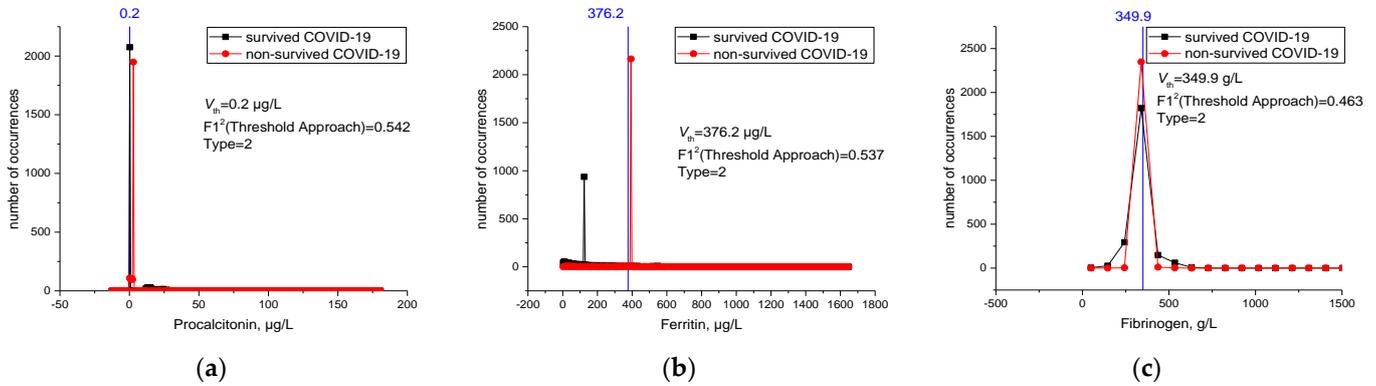

(a) (b) (c)

**Figure 9.** Histogram distributions and $F1^2$ results of (**a**) procalcitonin, (**b**) ferritin and (**c**) fibrinogen properties, according to the single-cut-off value approach in estimating COVID-19 mortality. $V_{th}$ (blue line) is the threshold for detecting COVID-19 mortality.

The amylase samples for the histogram distribution and classification results of the characteristics of surviving and non-surviving COVID-19 patients according to the two-threshold approach, are shown in Figure 10.

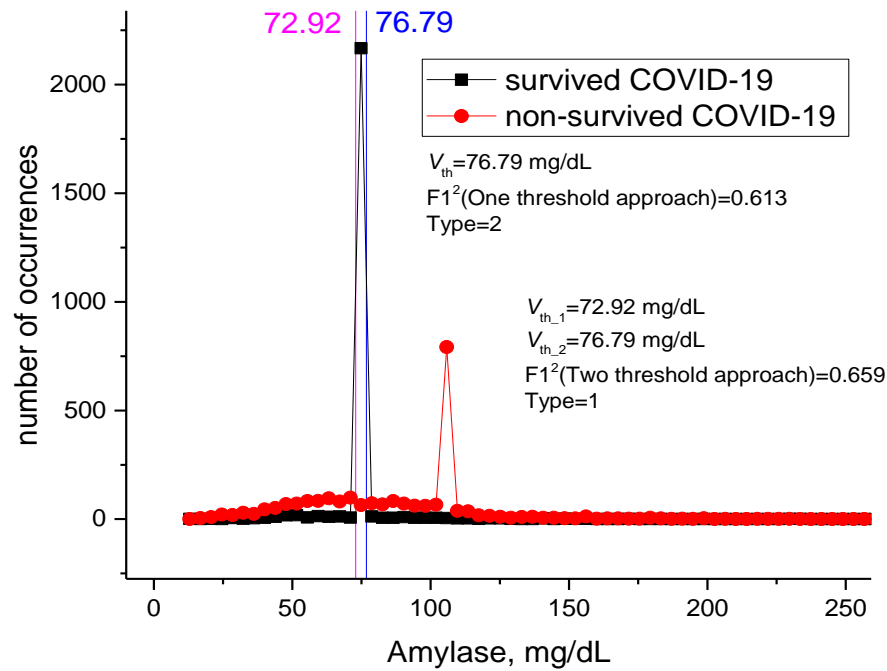

**Figure 10.** Histogram distributions and $F1^2$ results of amylase feature according to two-threshold value approach in estimating COVID-19 mortality. $V_{th\_1}$ (pink line) and $V_{th\_2}$ (blue line) is the threshold for detecting COVID-19 mortality.

3.4.2. Comparison of Spearman Correlation and HGB Model and Threshold Approach

It was observed that the performance of the HGB model with a single feature ($F1^2$) in the classification of patient groups who died from and survived COVID-19 was more successful than the classification made by considering one- and two-threshold values (Figure 11). In addition, Spearman's correlation gave a similar distribution of features in terms of importance, as the presented models.

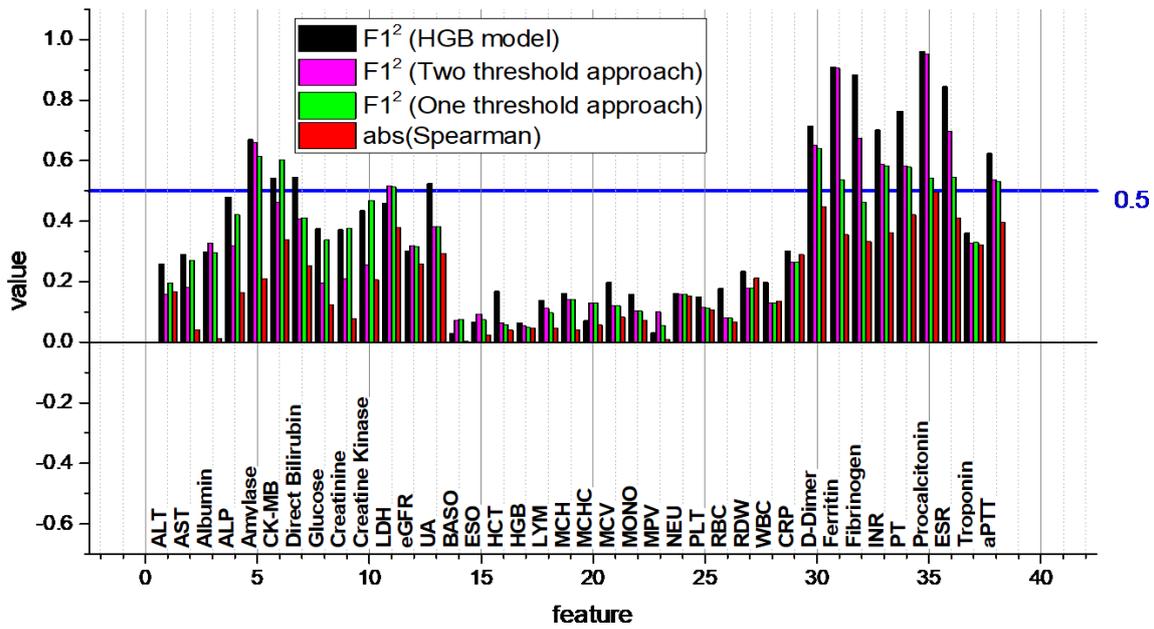

**Figure 11.** $F1^2$ metric of SARS-CoV-2-RBV3 dataset for different models.

*3.5. Investigation of the Effectiveness of the HGB Model Working on Two Features for the Detection of Surviving and Non-Surviving COVID-19*

For the detection of living and deceased COVID-19 patient-groups, the SMOTE-trained HGB model was run with dual features, and classification performances are presented in Table 4. In addition, $F1^2$ values related to binary properties and classification performances operated with the HGB model are visualized in two-dimensional space (Figure 12).

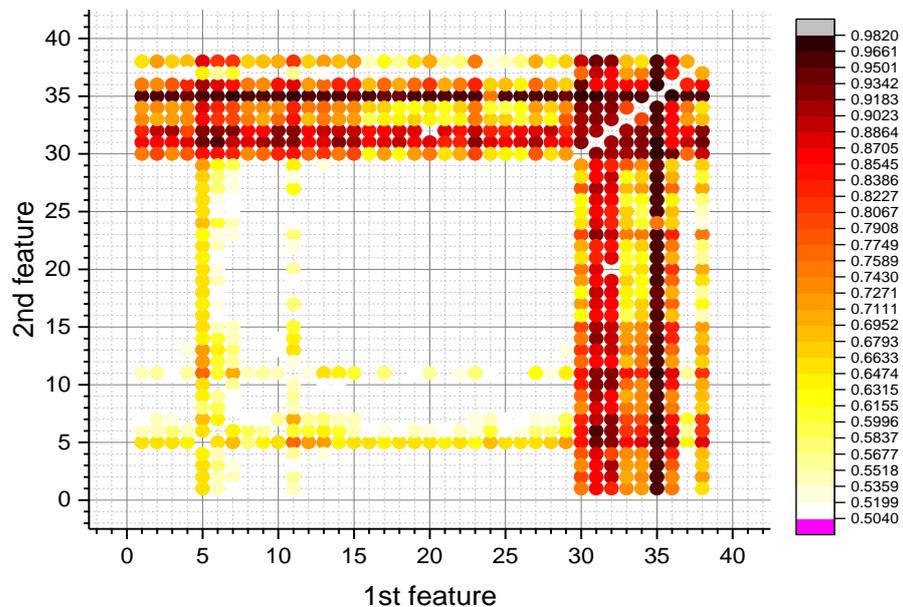

**Figure 12.** Feature pairs with the highest $F1^2$ value that was found with the HGB classifier for detection of surviving and non-surviving COVID-19 patients.

When Table 6 and Figure 12 are examined, the HGB model shows a classification performance of $F1^2 = 0.98$ with only D-dimer and PCT feature pair, in the detection of surviving and non-surviving patients. The classification performances of the feature pairs formed by PCT with ESR, D-Bil, Ferritin, and LDH were approximately $F1^2 = 0.98$. Both surviving and non-surviving patients with these feature pairs were identified, with high precision and recall values. PCT appears to be the feature that most closely matches other features in predicting disease mortality. After PCT, it can be said that ferritin is the most-matching property with other properties. In addition, $\geq 0.94$ $F1^2$ values were found in the patient-group classification of various feature pairs with the HGB model (Table 6)

In addition, according to the two-threshold approach, it was found that the majority of patients who died had fibrinogen values between 349.98 g/L and 379.05 g/L ($F1^2 = 0.68$), D-dimer values between 1009.99 µg/L and 10742.71 µg/L ($F1^2 = 0.65$), ESR values between 36.12 and 56.62 ($F1^2 = 0.70$), ferritin values between 376.2 µg/L and 396.0 µg/L ($F1^2 = 0.91$) and PCT values between 0.2 µg/L and 5.2 µg/L ($F1^2 = 0.95$) (Figure 8 and Table A3). It can be said that the determined value ranges of these features are the most important lethal-risk levels. It is noteworthy that procalcitonin and ferritin are the most important feature pairs in the detection of surviving and non-surviving COVID-19 patients, according to both the HGB classifier and the two-threshold approach.

**Table 6.** Feature pairs with the highest metrics found with the HGB classifier for detection of surviving and non-surviving COVID-19 patients.

| Feature Pairs | | | | Precision | | Recall | | F1 | | $F1^2$ |
|---|---|---|---|---|---|---|---|---|---|---|
| | | | | Surv. | Non-Surv. | Surv. | Non-Surv. | Surv. | Non-Surv. | |
| D-dimer | PCT | 30 | 35 | 0.9979 | 0.9867 | 0.9987 | 0.9786 | 0.9983 | 0.9825 | 0.98083 |
| PCT | ESR | 35 | 36 | 0.997 | 0.9911 | 0.9992 | 0.9704 | 0.9981 | 0.9805 | 0.97864 |
| D-Bil | PCT | 7 | 35 | 0.9987 | 0.9735 | 0.9975 | 0.9866 | 0.9981 | 0.9798 | 0.97794 |
| Ferritin | PCT | 31 | 35 | 0.997 | 0.9868 | 0.9987 | 0.9699 | 0.9979 | 0.9782 | 0.97615 |
| LDH | PCT | 11 | 35 | 0.9992 | 0.9648 | 0.9966 | 0.991 | 0.9979 | 0.9774 | 0.97535 |
| PT | PCT | 34 | 35 | 0.9953 | 0.9781 | 0.9979 | 0.9536 | 0.9966 | 0.9654 | 0.96212 |
| PCT | aPTT | 35 | 38 | 0.9975 | 0.9564 | 0.9958 | 0.975 | 0.9966 | 0.9643 | 0.96102 |
| CK-MB | PCT | 6 | 35 | 0.9983 | 0.9473 | 0.995 | 0.983 | 0.9966 | 0.9641 | 0.96082 |
| INR | PCT | 33 | 35 | 0.9941 | 0.9868 | 0.9987 | 0.9435 | 0.9964 | 0.9641 | 0.96063 |
| MCH | PCT | 19 | 35 | 0.9992 | 0.9386 | 0.9941 | 0.991 | 0.9966 | 0.9635 | 0.96022 |
| ALT | PCT | 1 | 35 | 0.9992 | 0.9386 | 0.9941 | 0.991 | 0.9966 | 0.9635 | 0.96022 |
| MCV | PCT | 21 | 35 | 0.9992 | 0.9386 | 0.9941 | 0.991 | 0.9966 | 0.9635 | 0.96022 |
| eGFR | PCT | 12 | 35 | 0.9992 | 0.9386 | 0.9941 | 0.991 | 0.9966 | 0.9635 | 0.96022 |
| Creatinine | PCT | 9 | 35 | 0.9992 | 0.9386 | 0.9941 | 0.991 | 0.9966 | 0.9635 | 0.96022 |
| RBC | PCT | 26 | 35 | 0.9992 | 0.9386 | 0.9941 | 0.991 | 0.9966 | 0.9635 | 0.96022 |
| Glucose | PCT | 8 | 35 | 0.9992 | 0.9386 | 0.9941 | 0.991 | 0.9966 | 0.9635 | 0.96022 |
| UA | PCT | 13 | 35 | 0.9992 | 0.9386 | 0.9941 | 0.991 | 0.9966 | 0.9635 | 0.96022 |
| WBC | PCT | 28 | 35 | 0.9987 | 0.9386 | 0.9941 | 0.9863 | 0.9964 | 0.9614 | 0.95794 |
| BASO | PCT | 14 | 35 | 0.9987 | 0.9386 | 0.9941 | 0.9865 | 0.9964 | 0.9613 | 0.95784 |
| PLT | PCT | 25 | 35 | 0.9987 | 0.9386 | 0.9941 | 0.9865 | 0.9964 | 0.9613 | 0.95784 |
| RDW | PCT | 27 | 35 | 0.9987 | 0.9386 | 0.9941 | 0.9865 | 0.9964 | 0.9613 | 0.95784 |
| AST | PCT | 2 | 35 | 0.9987 | 0.9386 | 0.9941 | 0.9865 | 0.9964 | 0.9613 | 0.95784 |
| PCT | Troponin | 35 | 37 | 0.9983 | 0.9386 | 0.9941 | 0.9821 | 0.9962 | 0.9593 | 0.95565 |
| CK | PCT | 10 | 35 | 0.9979 | 0.9431 | 0.9945 | 0.978 | 0.9962 | 0.9593 | 0.95565 |
| MPV | PCT | 23 | 35 | 0.9983 | 0.9386 | 0.9941 | 0.9817 | 0.9962 | 0.9593 | 0.95565 |
| MONO | PCT | 22 | 35 | 0.9983 | 0.9386 | 0.9941 | 0.9819 | 0.9962 | 0.9593 | 0.95565 |
| Albumin | PCT | 3 | 35 | 0.9992 | 0.9297 | 0.9933 | 0.9912 | 0.9962 | 0.958 | 0.95436 |
| MCHC | PCT | 20 | 35 | 0.9987 | 0.9298 | 0.9933 | 0.9878 | 0.996 | 0.9566 | 0.95277 |
| CK-MB | Ferritin | 6 | 31 | 0.9924 | 0.987 | 0.9987 | 0.9271 | 0.9955 | 0.9558 | 0.9515 |
| Amylase | PCT | 5 | 35 | 0.9966 | 0.9474 | 0.995 | 0.9648 | 0.9958 | 0.9554 | 0.95139 |
| HCT | PCT | 16 | 35 | 0.9979 | 0.9343 | 0.9937 | 0.9792 | 0.9958 | 0.9549 | 0.95089 |
| Ferritin | aPTT | 31 | 38 | 0.9915 | 0.9825 | 0.9983 | 0.9258 | 0.9949 | 0.9511 | 0.94625 |

| | | | | | | | | | |
|---|---|---|---|---|---|---|---|---|---|
| EOS | PCT | 15 | 35 | 0.997 | 0.9343 | 0.9937 | 0.9688 | 0.9954 | 0.9506 | 0.94623 |
| HGB | PCT | 17 | 35 | 0.9979 | 0.9253 | 0.9929 | 0.9768 | 0.9954 | 0.9495 | 0.94513 |
| LYM | PCT | 18 | 35 | 0.9962 | 0.9386 | 0.9941 | 0.9604 | 0.9951 | 0.9488 | 0.94415 |
| D-dimer | ESR | 30 | 36 | 0.9911 | 0.9825 | 0.9983 | 0.9161 | 0.9947 | 0.9477 | 0.94268 |
| CRP | PCT | 29 | 35 | 0.9958 | 0.9386 | 0.9941 | 0.9579 | 0.9949 | 0.9468 | 0.94197 |

*3.6. Concept of 1D and 2D Masks*

In order to understand the working principle of the HGB model in the classification of surviving and non-surviving COVID-19 patients, the cut-off values of one and two features and their sampling distributions were drawn, and the results were visualized with the masking technique.

3.6.1. 1D Mask of the HGB Model

Figure 13a shows the distribution of procalcitonin by patient groups on the original dataset, and shows how the patient groups were classified according to the threshold values of this feature. The procalcitonin value is used as an example for understanding the procedure for identifying patient groups according to a single feature of the HGB model. The classification results of the HGB model, which uses the cut-off values of procalcitonin to determine the patient groups, are visualized in the 1D-mask technique in Figure 13b.

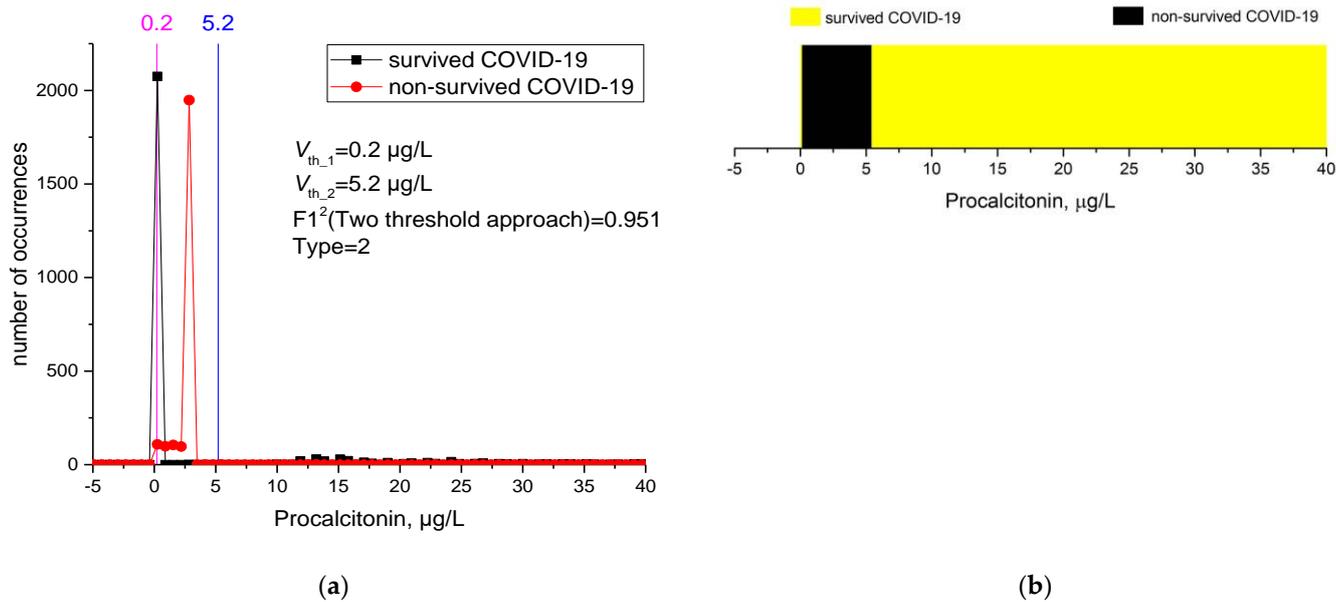

(**a**) (**b**)

**Figure 13.** (**a**) Distribution of the procalcitonin feature in the original data of patients who survived and those who died from COVID-19, and the two-threshold value for this feature in classification. (**b**) The 1D masking technique for classifying patient-groups in the HGB model operated with the procalcitonin feature.

3.6.2. 2D Mask of HGB Model

Figure 14a,c shows the distribution of D-dimer-ferritin and MCH-creatine kinase properties in two-dimensional space, according to patient groups on the original dataset. These feature pairs have been chosen as examples to understand the working principle of the HGB model with dual features. The results of the classification of living and deceased COVID-19 patients with the HGB model using these features were visualized with the 2D-masking technique (Figure 14b,d). D-dimer-ferritin properties were the feature pairs with the highest F1$^2$ score in the identification of patient groups with HGB, while MCH-Creatine kinase were the feature pairs with the lowest F1$^2$ score. Here, we have shown the working principle of these two contrasting features with HGB.

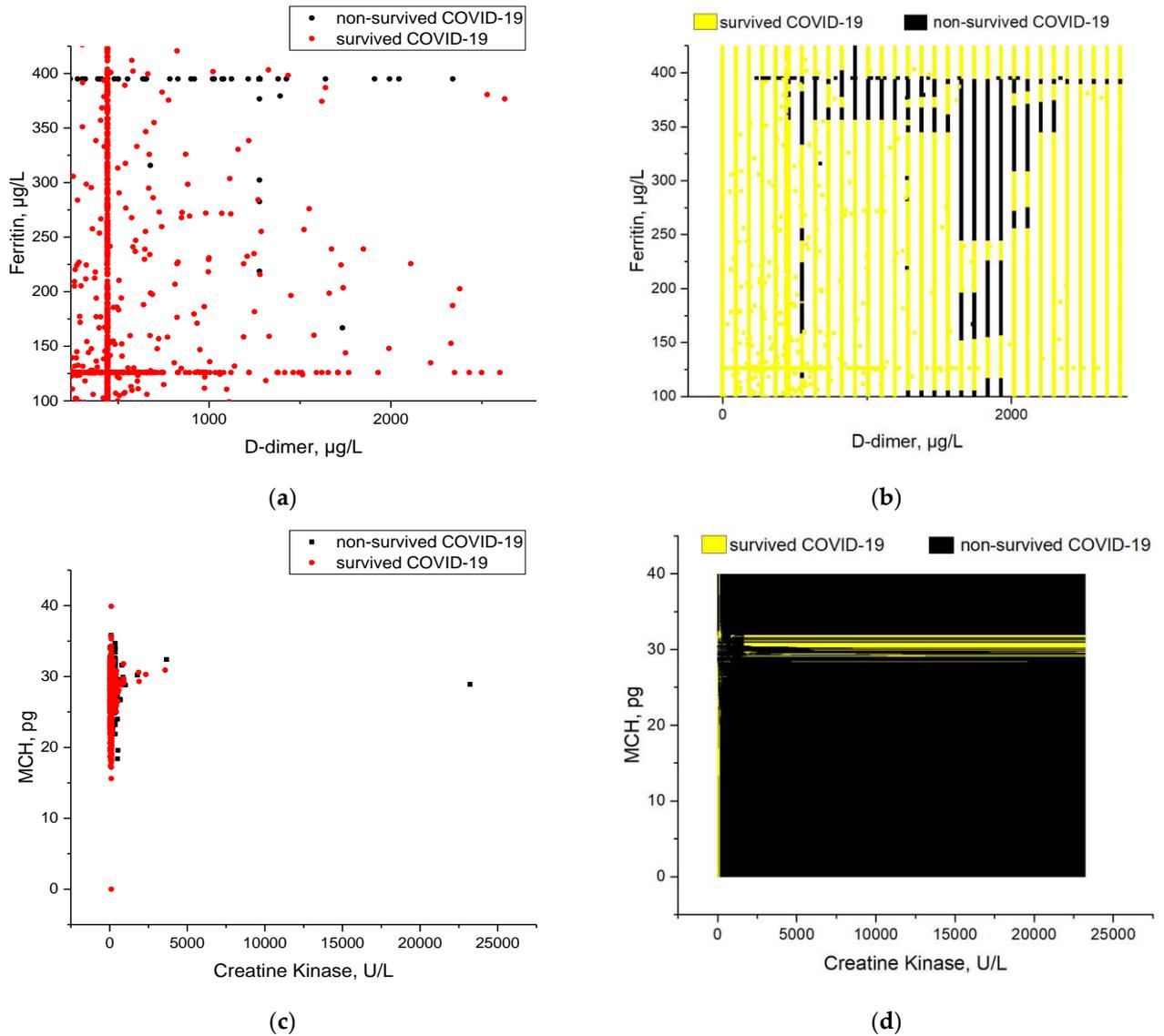

**Figure 14.** Distributions of non-surviving and surviving COVID-19 patients over the original data on D-dimer-ferritin (**a**) and CK-MCH (**c**) feature pairs. The 2D-masking technique for patient-group classification of the HGB model operated with D-dimer-ferritin (**b**) and CK-MCH (**d**) feature pairs.

## 4. Discussion

COVID-19, caused by the novel coronavirus SARS-CoV-2, is a new disease for humanity and contains many unknowns [16]. During the course of the disease, changes are observed in many biochemical parameters, as well as hematological abnormalities [1,4,15,23]

While most patients have mild symptoms, some patients may develop severe symptoms such as severe pneumonia, acute respiratory distress syndrome (ARDS), and multiple organ dysfunction syndromes (MODS) [4,23,46]. Therefore, early evaluation of patients who require special care, high mortality-expectation, and effective identification of relevant biomarkers on large sample groups are important to reduce mortality [4,13,29,35].

In this study, firstly, increasing and decreasing relationship-levels between living and deceased patient-groups and feature pairs were examined (Table 2). Then, 34 features were determined using a statistical approach to determine the most successful ML classifier-model in detecting living and deceased COVID-19 patients (Table 3). Our dataset was balanced with SMOTE, and our ML models were trained with the balanced dataset, as there was a large sample difference (91% versus 9%) between the groups of patients who lived and those who died from COVID-19, in our dataset. The patient groups were classified using 16 ML models operated with 34 features, and the most successful was the histogram-based gradient boosting (HGB) model ($F1^2 = 1$) (Table 4). Then, with the HGB model, the most important predictors (12 features) in estimating the mortality of COVID-19 were revealed, and lethal-risk factors of the disease were determined (Table 5). In addition, pairs of features with the highest classification-rate were determined by using binary combinations of all features to determine patient groups (Table 6). Moreover, classification results were found by calculating the most important cut-off values in the classification of patients who lived and those who died, according to one- and two-threshold values (Tables 5, A2 and A3).

In this study, patients who died and those who survived COVID-19 were highly associated with the feature pairs HGB-HCT, RBC-HCT, RBC-HGB, NEU-WBC, INR-PT (Figure 2b,c). We think that these pairs of features are associated with the prognosis of the disease and have significant negative effects on the immune system during the disease process. Moradi et al. [47] stated that the components of the immune system are the organs most frequently affected by COVID-19, after the lungs, and stated that necrosis and bleeding, as well as spleen atrophy and significant reductions in lymphocyte and neutrophil counts, may occur in these patients. In addition, Guzik et al. [48] noted that these features were highly correlated with the prognosis of the disease. Song et al. [49] determined that increased NEU, WBC, CRP, and D-dimer levels may reflect an imbalance in the inflammatory response, and these features can be considered as a possible indicator of disease severity in infectious diseases such as sepsis and bacteremia. In one study, it was reported that lower levels of RBC, lymphocytes, platelets, HGB, and higher neutrophils were observed in the peripheral blood system of severe COVID-19 patients [50].

In this study, there was a significant decrease in the level of the relationship between the patients who died and the pairs of albumin-glucose, ESR-D-dimer, creatinine-ALT, HCT-EOS, ESR-Fibrinogen, and ferritin-D-dimer properties when compared with the patients who survived (shown as "Down" in Table 2). Here, we can say that the applications applied to the patients who passed away have little effect on the values of these features, and that there are hidden relationship-structures between these feature pairs and mortality. We think that the decrease in the relationship structure between these feature pairs and the disease increases mortality. In addition, the relationship rate of all feature pairs shown as "Up" in Table 2 with deceased patients was significantly increased, compared with living patients. In particular, the greatly increased level of relationship between NEU-amylase, BASO-CKMB, MPV-AST, D-dimer-creatinine and MPV-UA feature pairs and patients who died made us think that important disorders such as kidney and liver functions occur in severe COVID-19 patients. Although the increasing relationship of these feature pairs with the patients who lost their lives points to the lack of self-care, we think that these feature pairs hide important information in the increasing mortality of the disease. It is understood that there are serious increases and decreases in the level of relationship between this feature and its various combinations in the period until death, in patients who lost their lives. We think that the difficulties in the management of this process and the serious changes in the levels of these feature pairs indicate very different complications in severe patients. In this context, we can say that the increase or decrease in one of these features has a significant effect on the metabolism of the other feature, depending on the severity of the disease.

Huyut et al. [12] stated that the patients who died had significant changes in liver- and kidney-function tests, cardiac-troponin and hemogram values, and parameters related to inflammation. They also stated that high ESR, PT, CRP, D-dimer, ferritin, and RDW values are the most effective predictors of mortality of COVID-19. Similarly, Chen et al. [14] and Tan et al. [51] determined that disorders resulting from hematological abnormalities were associated with disease severity. Many studies have reported that leukocytosis and lymphopenia levels are independent predictors of in-hospital mortality [12,46,49]. Huyut et al. [12] did not find EOS and other hematological values to be a predictive risk factor for COVID-19 mortality, while they stated that high NEU, WBC and RDW values are important mortality risk-indicators of COVID-19. Similarly, one study noted that neutrophils play an important role in inflammation, and this increase contributes to the development of ARDS [52]. In other studies, neutrophil was noted to be an independent predictor of severe disease, and associated with hypersensitivity pneumonia in SARS-CoV-2 [52,53]. Although some studies have indicated that increased amylase or lipase indicates pancreatic injury in COVID-19 patients, this has not been proven in other studies, and it has been stated that the increase in these enzymes can also be seen in other clinical conditions [54].

It is known that the use of the F1 score may be more useful than the accuracy value in cases where the data distributions are not equal. In the classification of surviving patients, all features were found to have a high F1 score for both the original and SMOTE-balanced datasets (Figure 4). In addition, in the classification of patients who died, it was found that the majority of the features had high F1 values for the SMOTE-balanced dataset, while this score was high for only some features in the original dataset (Figure 5).

To select the most important features in defining patient-groups, we defined an additional metric, ($F1^2$), equal to the product of the F1 metrics of the two classes (Equation (3)). We tested the HGB model on the original dataset for single (Figure 6 and Table 5) and dual features (Figure 12 and Table 6), although synthetic data on surviving and deceased patients were well predicted by the model (Table A1, Figures 4 and 5). PCT, ferritin, fibrinogen, ESR, PT, and D-dimer were found to be the most important features according to the $F1^2$ metric for the histogram-based gradient-boosting model operated with single features in the classification of surviving and deceased patient-groups (Figure 6 and Table 5). In addition, it was observed that PCT and ferritin were the most important feature-pairs in the identification of living and deceased patients (precision > 0.98, recall > 0.98, $F1^2$ > 0.98 in both living and deceased patients) (Table 6 and Figure 12). In addition, other feature pairs run with the HGB model produced an $F1^2$ value of ≥ 0.94 in identifying patient-groups (Table 6). Accordingly, our HGB model, which was trained with SMOTE, was found to largely accurately identify living and deceased COVID-19 patients. In addition, the performance of the HGB model with a single feature ($F1^2$) in the classification of patient groups was found to be more successful than the classification made by considering one and two cut-off-values (Figure 11). In addition, the approach of identifying patient groups based on the relationship structure (Spearman) of the characteristics of the patient groups produced the lowest $F1^2$ results (Figure 11).

In this study, in order to determine the critical-risk levels of the features in COVID-19 mortality, the lethal levels of the features were determined with one- and two-threshold approaches (see Section 2.4), and patient groups were classified according to these values [Figures 7 and 8 and Tables A2 and A3]. According to the one-threshold approach, PT values greater than 13.50 Sec ($F1^2$ = 0.58), D-dimer values greater than 1009.99 μg/L ($F1^2$ = 0.64), INR values greater than 1.51 ($F1^2$ = 0.58), amylase values greater than 76.79 mg/dL ($F1^2$ = 0.61) and CK-MB values greater than 18.86 U/L ($F1^2$ = 0.60) were found to be lethal critical-levels for COVID-19 mortality. According to the two-threshold approach, it was found that the majority of patients who died had fibrinogen values between 349.98 g/L and 379.05 g/L ($F1^2$ = 0.68), D-dimer values between 1009.99 μg/L and 10742.71 μg/L ($F1^2$ = 0.65), ESR values between 36.12 and 56.62 ($F1^2$ = 0.70), ferritin values between 376.2 μg/L and 396.0 μg/L ($F1^2$ = 0.91) and PCT values between 0.2 μg/L and 5.2 μg/L ($F1^2$ = 0.95). It can be said that the determined value ranges of these features are the most important lethal-risk levels. It is noteworthy that procalcitonin and ferritin are the most important feature pairs in the detection of surviving and non-surviving COVID-19 patients, according to both the HGB classifier and the two-threshold approach.

Similar to the findings in this study, many studies have supported the view that any significant increase in PCT levels reflects the development of a critical condition in COVID-19 [55–59]. Lima et al. [60] stated that, due to the characteristic structure of PCT in bacterial and viral infections, it may play a role in the prognosis of COVID-19. Ahmed et al. [55] noted that despite several limitations, elevated PCT levels can be used as a rapid indicator of criticality, a worsening clinical-picture, and even mortality, in COVID-19. Similarly, Lippi et al. [58] stated in a meta-analysis that procalcitonin levels above 0.5 µg/L were correlated with a 5-fold greater risk of serious infection in COVID-19 patients. In another study, Juneja et al. [61] showed that more than 96% of COVID-19 patients with low disease-severity had serum procalcitonin levels of less than 0.5 µg/L, and that these patients had better clinical outcomes. Additionally, Juneja et al. noted that PCT levels above 0.5 µg/L are associated with a more serious COVID-19 illness or secondary bacterial infection [61]. A meta-analysis involving Caucasians and South Asians found a strong association between PCT and the severity of COVID-19 [55]. This multiethnic assessment further reinforces the importance of PCT as a prognostic biomarker in cases of COVID-19. Lippi et al. [58], emphasizing the properties of PCT, its reliable kinetics and the potential relationship of its decreasing levels with infection resolution, stated that this feature may be a promising prognostic biomarker for COVID-19. These results support our results in our study.

In a meta-analysis examining a limited amount of the literature, Henry et al. [62] stated that high-hematological findings detected in COVID-19 patients and an increase in values such as D-dimer and IL-6 were accepted as an indicator of widespread cytokine-release. Similarly, Onur et al. [16] determined that the increase in biochemical parameters such as ferritin, fibrinogen, D-dimer, and troponin measured at the first hospitalization was associated with mortality. In other studies, Perricone et al. [18] and Torti et al. [63] noted that circulating ferritin levels may not only reflect the acute-phase response, but may also play a critical role in inflammation. In addition, some studies reported that ferritin as a signaling molecule may be a direct mediator of the immune system [17,64]. Similar to the ferritin findings in this study, Feld et al. [26] and Kernan and Carcillo. [65] stated that ferritin, the essential intracellular iron-storage protein, is an acute-phase reactant that is elevated in many inflammatory conditions, including acute infections. Onur et al. [16] stated that ferritin, an indicator of systemic inflammation, may be an indicator of disease severity and mortality. Winata and Kurniawan [66] emphasized that D-dimer and fibrinogen degradation product (FDP) are increased in all patients in the late stage of COVID-19. These results suggested that D-dimer and FDP levels were elevated due to increased hypoxia in severe COVID-19 patients, and that these properties were significantly associated with coagulation.

In addition, Huyut et al. [12], Mertoğlu et al. [23] and Huyut and İlkbahar [4] stated that increased fibrinogen, D-dimer, and CRP levels cause widespread inflammation and cytokine storm in severe COVID-19 patients, and they stated that high values of these features will increase mortality. In addition, high PT- and INR-values were interpreted as favoring hypercoagulation in a significant proportion of patients who died in this study. This result supported the idea that the risk of hypercoagulation is high in COVID-19 patients who die. In another study, similar to the findings in this study, increased PT, INR and low aPTT values were interpreted as favoring hypercoagulation in a significant proportion of patients who died [12]. These results contribute to the thought [12,21,23] that cardiovascular pathologies due to coagulation may be increased in patients who die.

## 5. Limitations of the Study

The data set in this article does not include the comorbidities of the patients and the inpatient/outpatient follow-up. However, in practice, it is seen that a training set collected during a certain time period cannot meet all these demands. In addition, this study was carried out only on the Turkish ethnicity. Results may need to be tested on other populations. However, the histogram-based gradient-boosting-model approach is easy to retrain and test with data from patients of different ethnicities. As more data becomes available, the algorithm will improve in terms of predictive performance of mortality from COVID-19.

## 6. Conclusions

In this study, the histogram-based gradient-boosting (HGB) model was the most successful ML classifier in detecting surviving and non-surviving COVID-19 patients ($F1^2$ = 1). Major changes were observed in many RBV values of patients who died from COVID-19. This situation indicated that self-care insufficiency developed due to the process in patients who died, but it also suggested that important disorders occurred in the functions of many organs such as the liver and kidney. In addition, we can say that an increase or decrease in an RBV value according to the severity of the disease, has a significant effect on the metabolism of another RBV value.

The HGB model, which was run with only procalcitonin and ferritin, correctly detected almost all of the COVID-19 patients, both living and deceased (precision > 0.98, recall > 0.98, $F1^2$ > 0.98). In addition, ferritin values between 376.2 µg/L and 396.0 µg/L ($F1^2$ = 0.91) and procalcitonin values between 0.2 µg/L and 5.2 µg/L ($F1^2$ = 0.95) were found to be fatal risk-levels for COVID-19.

In this study, we suggest that the HGB model and ferritin and procalcitonin properties can be used to obtain highly successful results in predicting the mortality of COVID-19. In addition, it was found to be remarkable that procalcitonin and ferritin were the most important features in the determination of patient groups, both with the HGB model and using the two-threshold approach. Accordingly, we think that the critical levels of ferritin and procalcitonin properties we have determined should be taken into account to reduce the lethality of the COVID-19 disease. These biomarkers and their critical levels can also serve as a risk-stratification tool for resource allocation and aggressive therapeutics, along with clinical details in over-crowded medical centers during the epidemic.

**Supplementary Materials:** The following supporting information can be downloaded at: www.mdpi.com/xxx/s1, SARS-CoV-2-RBV3_Dataset.zip. Kindly cite our paper when you wish to use this dataset.

**Author Contributions:** Conceptualization, M.T.H.; methodology, M.T.H., A.V. and M.B.; software, M.T.H., A.V. and M.B.; validation, M.T.H.; formal analysis, M.T.H.; investigation, M.T.H., A.V. and M.B.; resources, M.T.H.; data curation, M.T.H.; writing—original draft preparation, M.T.H., A.V. and M.B.; writing—review and editing, M.T.H., A.V. and M.B.; visualization M.T.H., A.V. and M.B.; supervision, M.T.H.; project administration, M.T.H.; funding acquisition, A.V. All authors have read and agreed to the published version of the manuscript.

**Funding:** This research was supported by the Russian Science Foundation (grant no. 22-11-00055, https://rscf.ru/en/project/22-11-00055/ (accessed on 22 June 2022)).

**Institutional Review Board Statement:** The dataset used in this study was collected in order to be used in various studies in the estimation of the diagnosis, prognosis and mortality of COVID-19. The necessary permissions for the collected dataset were given by the Ministry of Health of the Republic of Turkey and the Ethics Committee of Erzincan Binali Yıldırım University. This study was conducted in accordance with the 1989 Declaration of Helsinki. Erzincan Binali Yıldırım University Human Research Health and Sports Sciences Ethics Committee Decision Number: 2021/02-07.

**Informed Consent Statement:** In this study, a dataset including only routine blood-values, RT-PCR results (positive or negative) and treatment units of the patients was downloaded retrospectively from the information system of our hospital, in the digital environment. New samples were not taken from the patients. There is no information in the dataset that includes identifying characteristics of individuals. It was stated that the routine blood-values would only be used in academic studies, and written consent was obtained from the institutions for this. In addition, therefore, written informed consent was not administered to every patient.

**Data Availability Statement:** The data used in this study can be shared with the parties, provided that the article is cited.

**Acknowledgements:** We thank the method used by Erzincan Mengücek Gazi Training and Research Hospital for their support in obtaining the material used in this study. Special thanks to the editors of the journal and to the anonymous reviewers, for their constructive criticism and improvement suggestions.

**Conflicts of Interest:** The authors declare no conflicts of interest.

**Appendix A**

**Table A1.** The classification results of SARS-CoV-2-RBV3 dataset for the HGB model using a single-input feature. Classification metrics (Precision, Recall, F1, F1$^2$) separately for classes (survived COVID-19 and non-survived COVID-19).

| № | Feature | Precision | | Recall | | F1 | | F1$^2$ |
|---|---|---|---|---|---|---|---|---|
| | | Surv. | Non-Surv. | Surv. | Non-Surv. | Surv. | Non-Surv. | |
| 1 | ALT | 0.8558 | 0.3245 | 0.9292 | 0.1803 | 0.8909 | 0.2314 | 0.20615 |
| 2 | AST | 0.8904 | 0.3771 | 0.9369 | 0.2486 | 0.913 | 0.2981 | 0.27217 |
| 3 | Albumin | 0.6832 | 0.9341 | 0.9909 | 0.2216 | 0.8086 | 0.3581 | 0.28956 |
| 4 | ALP | 0.8794 | 0.623 | 0.9604 | 0.3331 | 0.918 | 0.4324 | 0.39694 |
| 5 | Amylase | 0.9112 | 0.9692 | 0.9968 | 0.5134 | 0.952 | 0.671 | 0.63879 |
| 6 | CK-MB | 0.8655 | 0.917 | 0.9909 | 0.3975 | 0.9239 | 0.554 | 0.51184 |
| 7 | D-Bil | 0.9602 | 0.5347 | 0.9554 | 0.5713 | 0.9578 | 0.5503 | 0.52708 |
| 8 | Glucose | 0.8583 | 0.535 | 0.9504 | 0.267 | 0.902 | 0.356 | 0.32111 |
| 9 | Creatinine | 0.8367 | 0.6227 | 0.9583 | 0.2699 | 0.8933 | 0.3763 | 0.33615 |
| 10 | CK | 0.8219 | 0.7678 | 0.9735 | 0.2964 | 0.8911 | 0.4267 | 0.38023 |
| 11 | LDH | 0.8498 | 0.8774 | 0.9867 | 0.3659 | 0.9126 | 0.5127 | 0.46789 |
| 12 | eGFR | 0.6849 | 0.9037 | 0.9868 | 0.2174 | 0.8082 | 0.3501 | 0.28295 |
| 13 | UA | 0.8824 | 0.7586 | 0.9743 | 0.3858 | 0.926 | 0.5106 | 0.47282 |
| 14 | BASO | 0.9953 | 0.0088 | 0.9124 | 0.0786 | 0.952 | 0.0157 | 0.01495 |
| 15 | EOS | 0.9818 | 0.013 | 0.9116 | 0.0533 | 0.9454 | 0.0208 | 0.01966 |
| 16 | HCT | 0.9057 | 0.0967 | 0.9123 | 0.0853 | 0.9088 | 0.0887 | 0.08061 |
| 17 | HGB | 0.9873 | 0.0264 | 0.9131 | 0.1552 | 0.9488 | 0.045 | 0.0427 |
| 18 | LYM | 0.8481 | 0.1971 | 0.9165 | 0.1071 | 0.8808 | 0.1382 | 0.12173 |
| 19 | MCH | 0.9543 | 0.1097 | 0.9175 | 0.1955 | 0.9355 | 0.1379 | 0.12901 |
| 20 | MCHC | 0.9797 | 0.0439 | 0.914 | 0.1375 | 0.9457 | 0.0663 | 0.0627 |
| 21 | MCV | 0.8079 | 0.2544 | 0.9182 | 0.115 | 0.8594 | 0.1581 | 0.13587 |
| 22 | MONO | 0.9061 | 0.1665 | 0.9185 | 0.1491 | 0.9122 | 0.1569 | 0.14312 |
| 23 | MPV | 0.9949 | 0.0043 | 0.912 | 0.1 | 0.9516 | 0.0083 | 0.0079 |
| 24 | NEU | 0.4962 | 0.6972 | 0.9442 | 0.1181 | 0.6503 | 0.2019 | 0.1313 |
| 25 | PLT | 0.8697 | 0.1671 | 0.9154 | 0.1103 | 0.8918 | 0.1319 | 0.11763 |
| 26 | RBC | 0.8837 | 0.1185 | 0.9122 | 0.0888 | 0.8976 | 0.1007 | 0.09039 |
| 27 | RDW | 0.9082 | 0.2454 | 0.9258 | 0.2067 | 0.9169 | 0.2241 | 0.20548 |
| 28 | WBC | 0.9154 | 0.1798 | 0.9206 | 0.1629 | 0.9178 | 0.1678 | 0.15401 |
| 29 | CRP | 0.7386 | 0.6888 | 0.961 | 0.2034 | 0.835 | 0.3136 | 0.26186 |
| 30 | D-dimer | 0.937 | 0.842 | 0.984 | 0.5677 | 0.9599 | 0.6769 | 0.64976 |
| 31 | Ferritin | 0.9852 | 0.9253 | 0.9928 | 0.8655 | 0.9889 | 0.8915 | 0.8816 |
| 32 | Fibrinogen | 0.9805 | 0.9562 | 0.9957 | 0.8274 | 0.9881 | 0.8863 | 0.87575 |
| 33 | INR | 0.9315 | 0.847 | 0.9844 | 0.548 | 0.9571 | 0.663 | 0.63456 |
| 34 | PT | 0.959 | 0.8253 | 0.9828 | 0.6588 | 0.9707 | 0.7312 | 0.70978 |
| 35 | Procalcitonin | 0.9992 | 0.9386 | 0.9941 | 0.991 | 0.9966 | 0.9635 | 0.96022 |
| 36 | ESR | 0.967 | 0.8684 | 0.9871 | 0.7229 | 0.9769 | 0.7868 | 0.76862 |
| 37 | Troponin | 0.967 | 0.2898 | 0.9339 | 0.4699 | 0.9501 | 0.3547 | 0.337 |
| 38 | aPTT | 0.8837 | 0.8862 | 0.9878 | 0.4266 | 0.9327 | 0.5747 | 0.53602 |

Table A2. Results of classification according to the one-threshold approach of surviving and non-surviving COVID-19 patients, for the balanced dataset.

| № | Feature (Units) | Type | $V_{th}$ | $A_{th}$ | Precision | | Recall | | F1 | | F1² |
|---|---|---|---|---|---|---|---|---|---|---|---|
| | | | | | Surv. | Non-Surv. | Surv. | Non-Surv. | Surv. | Non-Surv. | |
| 1 | ALT (U/L) | 1 | 34.84 | 0.649 | 0.648 | 0.665 | 0.952 | 0.157 | 0.771 | 0.254 | 0.19583 |
| 2 | AST (U/L) | 2 | 33.472 | 0.806 | 0.839 | 0.476 | 0.942 | 0.226 | 0.887 | 0.306 | 0.27142 |
| 3 | Albumin (g/L) | 2 | 49.08 | 0.918 | 0.988 | 0.206 | 0.927 | 0.632 | 0.956 | 0.311 | 0.29732 |
| 4 | ALP (U/L) | 2 | 85.305 | 0.868 | 0.893 | 0.618 | 0.96 | 0.362 | 0.925 | 0.456 | 0.4218 |
| 5 | Amylase (U/L) | 2 | 76.79 | 0.936 | 0.966 | 0.627 | 0.963 | 0.646 | 0.965 | 0.636 | 0.61374 |
| 6 | CK-MB (U/L) | 2 | 18.86 | 0.92 | 0.935 | 0.764 | 0.976 | 0.538 | 0.955 | 0.631 | 0.6026 |
| 7 | D-Bil. (mg/dL) | 2 | 0.12985 | 0.842 | 0.854 | 0.725 | 0.969 | 0.328 | 0.908 | 0.452 | 0.41042 |
| 8 | Glucose (mg/dL) | 2 | 136.854 | 0.834 | 0.862 | 0.554 | 0.951 | 0.283 | 0.904 | 0.374 | 0.3381 |
| 9 | Creatinine (mg/dL) | 2 | 1.16656 | 0.877 | 0.918 | 0.464 | 0.946 | 0.358 | 0.932 | 0.404 | 0.37653 |
| 10 | CK (U/L) | 2 | 116.1 | 0.887 | 0.912 | 0.631 | 0.962 | 0.414 | 0.936 | 0.5 | 0.468 |
| 11 | LDH (U/L) | 2 | 253.26 | 0.874 | 0.875 | 0.867 | 0.985 | 0.406 | 0.927 | 0.553 | 0.51263 |
| 12 | eGFR | 1 | 82.57429 | 0.77 | 0.772 | 0.751 | 0.969 | 0.245 | 0.859 | 0.369 | 0.31697 |
| 13 | UA (mg/dL) | 2 | 39.01 | 0.818 | 0.824 | 0.755 | 0.972 | 0.298 | 0.892 | 0.427 | 0.38088 |
| 14 | BASO (10³/μL) | 2 | 0.01026 | 0.36 | 0.331 | 0.657 | 0.907 | 0.088 | 0.485 | 0.155 | 0.07517 |
| 15 | EOS (10³/μL) | 2 | 0.01323 | 0.368 | 0.344 | 0.614 | 0.9 | 0.084 | 0.498 | 0.148 | 0.0737 |
| 16 | HCT (%) | 1 | 44.0946 | 0.261 | 0.203 | 0.854 | 0.934 | 0.096 | 0.334 | 0.172 | 0.05745 |
| 17 | HGB (g/L) | 1 | 15.3972 | 0.229 | 0.162 | 0.906 | 0.946 | 0.096 | 0.277 | 0.174 | 0.0482 |
| 18 | LYM (10³/μL) | 1 | 1.72672 | 0.414 | 0.384 | 0.712 | 0.931 | 0.102 | 0.544 | 0.179 | 0.09738 |
| 19 | MCH (pg) | 2 | 29.6058 | 0.721 | 0.761 | 0.318 | 0.919 | 0.116 | 0.832 | 0.17 | 0.14144 |
| 20 | MCHC (g/dL) | 1 | 33.696 | 0.56 | 0.56 | 0.558 | 0.928 | 0.111 | 0.699 | 0.185 | 0.12931 |
| 21 | MCV (fL) | 2 | 83.7456 | 0.503 | 0.489 | 0.639 | 0.932 | 0.11 | 0.642 | 0.187 | 0.12005 |
| 22 | MONO (10³/μL) | 2 | 0.45078 | 0.422 | 0.392 | 0.73 | 0.936 | 0.106 | 0.553 | 0.185 | 0.10231 |
| 23 | MPV (fL) | 1 | 11.0988 | 0.265 | 0.214 | 0.785 | 0.91 | 0.09 | 0.346 | 0.161 | 0.05571 |
| 24 | NEU (10³/μL) | 2 | 4.379 | 0.571 | 0.56 | 0.691 | 0.948 | 0.134 | 0.704 | 0.224 | 0.1577 |
| 25 | PLT (10³/μL) | 1 | 245.85 | 0.451 | 0.423 | 0.73 | 0.941 | 0.111 | 0.584 | 0.193 | 0.11271 |
| 26 | RBC (10⁶/μL) | 1 | 5.06844 | 0.34 | 0.294 | 0.803 | 0.938 | 0.101 | 0.448 | 0.179 | 0.08019 |
| 27 | RDW (%) | 2 | 13.2096 | 0.598 | 0.585 | 0.73 | 0.956 | 0.148 | 0.726 | 0.246 | 0.1786 |
| 28 | WBC (10³/μL) | 2 | 6.2006 | 0.492 | 0.468 | 0.738 | 0.948 | 0.12 | 0.626 | 0.207 | 0.12958 |
| 29 | CRP (mg/L) | 2 | 19.488 | 0.72 | 0.719 | 0.738 | 0.965 | 0.205 | 0.824 | 0.321 | 0.2645 |
| 30 | D-dimer (μg/L) | 2 | 1009.998 | 0.92 | 0.922 | 0.906 | 0.99 | 0.533 | 0.955 | 0.671 | 0.6408 |
| 31 | Ferritin (μg/L) | 2 | 376.2 | 0.878 | 0.871 | 0.94 | 0.993 | 0.419 | 0.928 | 0.579 | 0.53731 |
| 32 | Fibrinogen (mg/dL) | 2 | 349.98608 | 0.834 | 0.82 | 0.979 | 0.997 | 0.349 | 0.9 | 0.515 | 0.4635 |
| 33 | INR | 2 | 1.15151 | 0.909 | 0.918 | 0.811 | 0.98 | 0.495 | 0.948 | 0.615 | 0.58302 |
| 34 | PT (Sec) | 2 | 13.50512 | 0.901 | 0.903 | 0.88 | 0.987 | 0.471 | 0.943 | 0.614 | 0.579 |
| 35 | PCT (ng/mL) | 2 | 0.2 | 0.882 | 0.878 | 0.923 | 0.991 | 0.427 | 0.931 | 0.583 | 0.54277 |
| 36 | ESR (nm/hr) | 2 | 36.125 | 0.883 | 0.88 | 0.918 | 0.991 | 0.43 | 0.932 | 0.585 | 0.54522 |
| 37 | Troponin (ng/L) | 2 | 13.2 | 0.906 | 0.968 | 0.279 | 0.932 | 0.461 | 0.949 | 0.348 | 0.33025 |
| 38 | aPTT (Sec) | 1 | 32.4594 | 0.875 | 0.87 | 0.931 | 0.992 | 0.413 | 0.927 | 0.573 | 0.53117 |

Table A3. Results of classification according to the two-threshold approach of surviving and non-surviving COVID-19 patients, for the balanced dataset.

| № | Feature (Units) | Type | $V_{th\_1}$ | $V_{th\_2}$ | $A_{th}$ | Precision Surv. | Precision Non-Surv. | Recall Surv. | Recall Non-Surv. | F1 Surv. | F1 Non-Surv. | F1 Surv. | $F1^2$ |
|---|---|---|---|---|---|---|---|---|---|---|---|---|---|
| 1 | ALT (U/L) | 1 | 34.84 | 35.36 | 0.518 | 0.483 | 0.876 | 0.975 | 0.143 | 0.646 | 0.246 | 0.15892 | |
| 2 | AST (U/L) | 1 | 32.949 | 33.472 | 0.536 | 0.492 | 0.979 | 0.996 | 0.16 | 0.659 | 0.275 | 0.18123 | |
| 3 | Albumin (g/L) | 1 | 36.81 | 49.08 | 0.808 | 0.826 | 0.627 | 0.957 | 0.262 | 0.887 | 0.37 | 0.32819 | |
| 4 | ALP (U/L) | 1 | 83.582 | 85.305 | 0.725 | 0.701 | 0.97 | 0.996 | 0.242 | 0.823 | 0.388 | 0.31932 | |
| 5 | Amylase (U/L) | 1 | 72.92 | 76.79 | 0.922 | 0.917 | 0.974 | 0.997 | 0.535 | 0.955 | 0.691 | 0.6599 | |
| 6 | CK-MB (U/L) | 1 | 18.4 | 18.86 | 0.832 | 0.816 | 0.996 | 0.999 | 0.347 | 0.898 | 0.515 | 0.46247 | |
| 7 | D-Bil. (mg/dL) | 1 | 0.04995 | 0.12985 | 0.836 | 0.845 | 0.747 | 0.971 | 0.322 | 0.904 | 0.45 | 0.4068 | |
| 8 | Glucose (mg/dL) | 1 | 135.631 | 136.854 | 0.557 | 0.514 | 0.991 | 0.998 | 0.167 | 0.679 | 0.286 | 0.19419 | |
| 9 | Creatinine (mg/dL) | 1 | 0.96492 | 1.16656 | 0.617 | 0.595 | 0.845 | 0.975 | 0.171 | 0.739 | 0.284 | 0.20988 | |
| 10 | CK (U/L) | 1 | 92.88 | 116.1 | 0.663 | 0.636 | 0.931 | 0.989 | 0.201 | 0.774 | 0.331 | 0.25619 | |
| 11 | LDH (U/L) | 2 | 253.26 | 597.64 | 0.876 | 0.877 | 0.867 | 0.985 | 0.411 | 0.928 | 0.557 | 0.5169 | |
| 12 | eGFR | 1 | 82.5742 | 146.2250 | 0.77 | 0.772 | 0.755 | 0.97 | 0.246 | 0.859 | 0.371 | 0.31869 | |
| 13 | UA (mg/dL) | 1 | 0 | 39.01 | 0.818 | 0.824 | 0.755 | 0.972 | 0.298 | 0.892 | 0.427 | 0.38088 | |
| 14 | BASO ($10^3/\mu L$) | 1 | 0.00988 | 0.01026 | 0.322 | 0.277 | 0.777 | 0.926 | 0.096 | 0.426 | 0.17 | 0.07242 | |
| 15 | EOS ($10^3/\mu L$) | 2 | 0.01323 | 0.11907 | 0.574 | 0.596 | 0.352 | 0.903 | 0.079 | 0.718 | 0.129 | 0.09262 | |
| 16 | HCT (%) | 2 | 30.1257 | 44.0946 | 0.293 | 0.246 | 0.768 | 0.915 | 0.091 | 0.388 | 0.163 | 0.06324 | |
| 17 | HGB (g/L) | 2 | 9.5128 | 15.3972 | 0.252 | 0.193 | 0.85 | 0.929 | 0.094 | 0.319 | 0.169 | 0.05391 | |
| 18 | LYM ($10^3/\mu L$) | 2 | 0.59356 | 1.72672 | 0.481 | 0.466 | 0.635 | 0.928 | 0.105 | 0.62 | 0.18 | 0.1116 | |
| 19 | MCH (pg) | 2 | 29.6058 | 35.6706 | 0.722 | 0.762 | 0.313 | 0.918 | 0.115 | 0.833 | 0.168 | 0.13994 | |
| 20 | MCHC (g/dL) | 2 | 28.431 | 33.696 | 0.562 | 0.563 | 0.558 | 0.928 | 0.112 | 0.701 | 0.186 | 0.13039 | |
| 21 | MCV (fL) | 2 | 83.7456 | 113.0624 | 0.503 | 0.489 | 0.639 | 0.932 | 0.11 | 0.642 | 0.188 | 0.1207 | |
| 22 | MONO ($10^3/\mu L$) | 2 | 0.45078 | 6.70023 | 0.423 | 0.393 | 0.73 | 0.936 | 0.106 | 0.554 | 0.185 | 0.10249 | |
| 23 | MPV (fL) | 2 | 9.9018 | 11.0988 | 0.539 | 0.549 | 0.438 | 0.908 | 0.087 | 0.685 | 0.146 | 0.10001 | |
| 24 | NEU ($10^3/\mu L$) | 2 | 4.379 | 24.853 | 0.573 | 0.561 | 0.691 | 0.948 | 0.134 | 0.705 | 0.225 | 0.15862 | |
| 25 | PLT ($10^3/\mu L$) | 2 | 108.025 | 245.85 | 0.474 | 0.453 | 0.687 | 0.936 | 0.11 | 0.611 | 0.19 | 0.11609 | |
| 26 | RBC ($10^6/\mu L$) | 2 | 0.00722 | 5.06844 | 0.34 | 0.295 | 0.803 | 0.938 | 0.101 | 0.449 | 0.179 | 0.08037 | |
| 27 | RDW (%) | 2 | 13.2096 | 21.1712 | 0.603 | 0.592 | 0.717 | 0.955 | 0.148 | 0.731 | 0.245 | 0.1791 | |
| 28 | WBC ($10^3/\mu L$) | 2 | 6.2006 | 44.054 | 0.493 | 0.469 | 0.738 | 0.948 | 0.12 | 0.627 | 0.207 | 0.12979 | |
| 29 | CRP (mg/L) | 2 | 19.488 | 252.938 | 0.722 | 0.721 | 0.73 | 0.964 | 0.205 | 0.825 | 0.32 | 0.264 | |
| 30 | D-dimer (µg/L) | 2 | 1009.99 | 10742.70 | 0.923 | 0.925 | 0.906 | 0.99 | 0.544 | 0.956 | 0.68 | 0.65008 | |
| 31 | Ferritin (µg/L) | 2 | 376.2 | 396 | 0.984 | 0.989 | 0.931 | 0.993 | 0.897 | 0.991 | 0.914 | 0.90577 | |
| 32 | Fibrinogen (mg/dL) | 2 | 349.986 | 379.054 | 0.927 | 0.923 | 0.97 | 0.997 | 0.553 | 0.958 | 0.704 | 0.67443 | |
| 33 | INR | 2 | 1.15151 | 10.4753 | 0.91 | 0.92 | 0.811 | 0.98 | 0.5 | 0.949 | 0.619 | 0.58743 | |
| 34 | PT (Sec) | 2 | 13.5051 | 110.0950 | 0.902 | 0.904 | 0.88 | 0.987 | 0.475 | 0.944 | 0.617 | 0.58245 | |
| 35 | PCT (ng/mL) | 2 | 0.2 | 5.2 | 0.992 | 1 | 0.918 | 0.992 | 0.995 | 0.996 | 0.955 | 0.95118 | |
| 36 | ESR (nm/hr) | 2 | 36.125 | 56.625 | 0.939 | 0.944 | 0.888 | 0.988 | 0.609 | 0.966 | 0.723 | 0.69842 | |
| 37 | Troponin (ng/L) | 2 | 13.2 | 3269.2 | 0.906 | 0.969 | 0.275 | 0.931 | 0.464 | 0.95 | 0.345 | 0.32775 | |
| 38 | aPTT (Sec) | 2 | 22.1582 | 32.4594 | 0.878 | 0.873 | 0.927 | 0.992 | 0.419 | 0.929 | 0.577 | 0.53603 | |